\documentclass[letterpaper, 10 pt, journal, twoside]{IEEEtran}
\usepackage{amsmath}
\usepackage{amsfonts}
\usepackage{algorithmic}
\usepackage{algorithm}
\usepackage{array}
\usepackage[caption=false,font=normalsize,labelfont=sf,textfont=sf]{subfig}
\usepackage{textcomp}
\usepackage{stfloats}
\usepackage{url}
\usepackage{verbatim}
\usepackage{graphicx}
\usepackage{cite}
\usepackage{bbding}
\usepackage{siunitx}
\usepackage{bm}
\usepackage{color}
\usepackage{xcolor}
\newcommand{\red}[1]{\textcolor{black}{#1}}
\newcommand{\rred}[1]{\textcolor{black}{#1}}

\begin{document}

\title{{AdapJ: An Adaptive Extended Jacobian Controller for Soft Manipulators}}
\author{Zixi Chen, Xuyang Ren, Yuya Hamamatsu, Gastone Ciuti, Donato Romano, and Cesare Stefanini
\thanks{This work was supported by the European Union by the Next Generation EU project ECS00000017 ‘Ecosistema dell’Innovazione’ Tuscany Health Ecosystem (THE, PNRR, Spoke 9: Robotics and Automation for Health), the Italian Space Agency (ASI) DC-DSR-UVS-2022-375 Project “pRomoting pEdogenesis throuGh lunar sOil-terrestriaL organIsms interaction For moon Fertilization – REGOLIFE” [ASI N.: 2024-7-U.0; CUP: J83C24000310005], and BRIEF “Biorobotics Research and Innovation Engineering Facilities” (project identification code IR0000036), project funded under the National Recovery and Resilience Plan (NRRP), Mission 4 Component 2 Investment 3.1 of Italian Ministry of University and Research funded by the European Union – NextGenerationEU. 
Z. Chen and X. Ren contributed equally to this work. \emph{Corresponding authors: Xuyang Ren.}}
\thanks{Z. Chen, G. Ciuti, D. Romano, and C. Stefanini are with the Biorobotics Institute and the Department of Excellence in Robotics and AI, Scuola Superiore Sant’Anna, 56127 Pisa, Italy. (email: zixi.chen@santannapisa.it; gastone.ciuti@santannapisa.it; donato.romano@santannapisa.it; cesare.stefanini@santannapisa.it)}
\thanks{X. Ren is with the Multi-scale Medical Robotics Centre and Chow Yuk Ho Technology Centre for Innovative Medicine, The Chinese University of Hong Kong, Hong Kong, China. (email: xuyang.ren.cn@gmail.com)}
\thanks{Y. Hamamatsu is with the Department of computer systems, Tallinn University of Technology, Tallinn, Estonia. (email: yuya.hamamatsu@taltech.ee)}
}


\maketitle
\begin{abstract}
The nonlinearity and hysteresis of soft robot motions present challenges for control.
To solve these issues, the Jacobian controller has been applied to approximate the nonlinear behaviors in a linear format.
Accurate controllers like neural networks can handle delayed and nonlinear motions, but they require large datasets and exhibit low adaptability.
\red{Based on a novel analysis on these controllers}, we propose an adaptive extended Jacobian controller, \textit{AdapJ}, for soft manipulators. 
This controller retains the concise format of the Jacobian controller but introduces independent parameters.
Similar to neural networks, its initialization and updating mechanism leverages the inverse model without building the corresponding forward model.
In the experiments, we first compare the performance of the Jacobian controller, model predictive controller, neural network controller, \red{iterative feedback controller}, and AdapJ in simulation. 
We further analyze how AdapJ parameters adapt in response to the physical property change.
Then, real-world experiments have validated that AdapJ outperforms \red{the neural network controller, model predictive controller, and iterative feedback controller} with fewer training samples and adapts robustly to varying conditions, including different control frequencies, material softness, and external disturbances.
Future work may include online adjustment of the controller format and adaptability validation in more scenarios.
\end{abstract}


\section{Introduction}
\label{sec1}
\IEEEPARstart{L}{everaging} the soft materials and flexible structures, soft robots offer higher degrees of freedom than traditional rigid robots.
As a result, soft robot manipulators have been utilized in many applications, such as minimally invasive surgery \cite{11ST} and exploration in confined spaces \cite{14MY}.
Their inherent compliance also reduces the risk of damaging objects, the surrounding environment, or human collaborators during motion and manipulation.
For example, soft manipulators can be applied for human-robot interaction and demonstrations \cite{14JQ}, and some can even be deployed as assistive robots for elderly individuals during showering activities \cite{23ZT}. 
Complex soft manipulators, like modular soft robots, can perform challenging tasks like shape control \cite{24ZCd} and high-level manipulation \cite{21HJ} utilizing dedicated control strategies.
Overall, soft manipulators represent a critical component of soft robotics, offering distinct advantages such as safety and adaptability, and have garnered significant attention from researchers across a wide range of disciplines.

\begin{figure}[t]
\centering
\includegraphics[width=\linewidth]{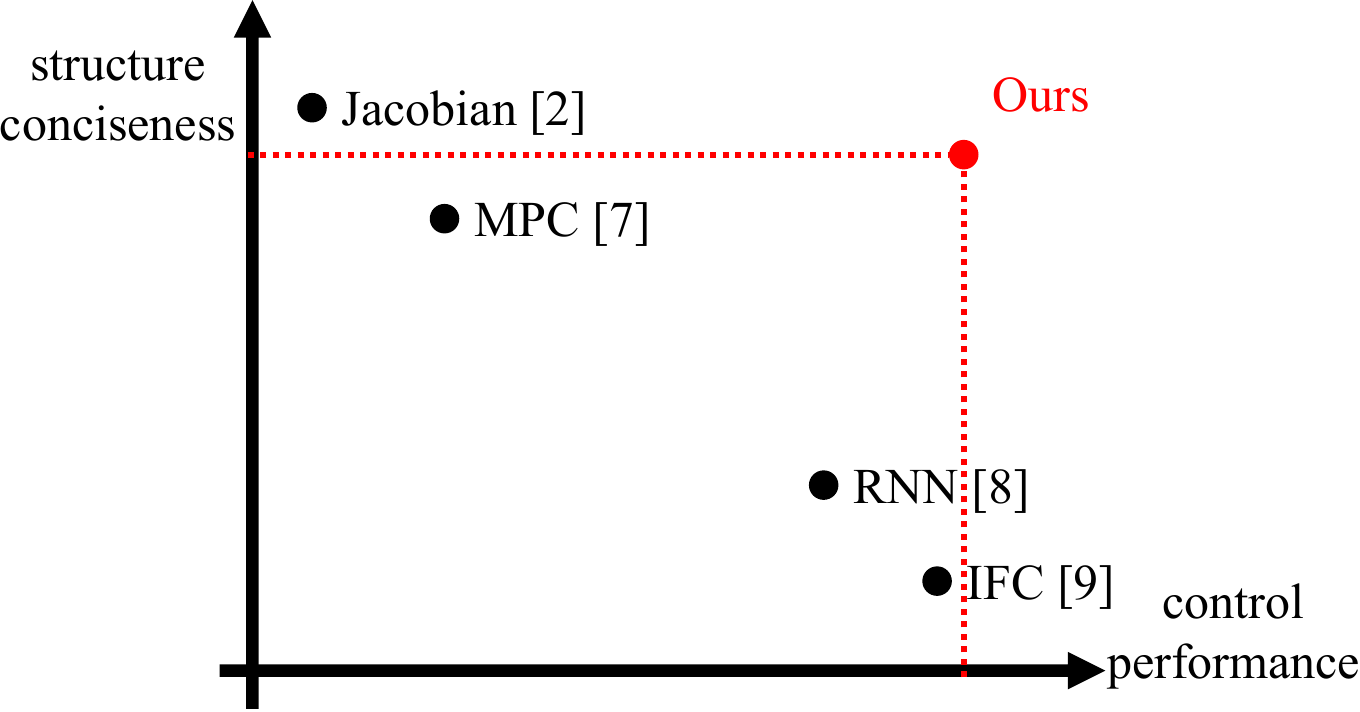}
\caption{
\red{Qualitative comparison of soft robot controllers. The proposed controller maintains a low structural complexity and computational cost, only slightly greater than that of the Jacobian controller, while outperforming the Jacobian controller \cite{14MY}, model predictive controller \cite{19ZT}, RNN controller \cite{23ZCc}, and iterative feedback controller \cite{24XL} in terms of control performance considering accuracy and adaptability.}
}
\label{fig1.1}
\end{figure}

It is vital to propose suitable controllers for soft manipulators to achieve challenging tasks. 
These controllers should be able to cope with the nonlinearity and hysteresis of soft robotic systems, and plenty of controllers have been introduced to address these challenges \cite{24ZCa}. 
For instance, Jacobian approaches are intuitive and straightforward to implement.
Although the linear and transient nature does not inherently accommodate the nonlinearity and hysteresis of soft robot motions, high-frequency online updates of the Jacobian matrix are applied to mitigate this issue.
Controllers based on physical models like Piecewise Constant Curvature (PCC) \cite{18CS} and Cosserat rods \cite{18FR} are explainable, but they are sophisticated and rely on theoretical assumptions that may not hold in practice.
Discrepancies between the actual behavior of soft robots and their idealized models caused by complex material properties and unpredictable deformations can lead to control inaccuracies.
Statistical controllers, including the Gaussian mixture model \cite{19BY} and the Gaussian process regression \cite{19GF}, rely on statistical models, which require a moderate amount of training data and yield moderate performance.
Neural networks (NNs), especially recurrent neural networks (RNNs) \cite{22DW}, have gained popularity in soft robot control due to their nonlinear activation functions and effective network structures.
Trained NNs can serve as global controllers across the entire workspace; however, their large parameter sets make real-time online updates challenging, limiting their adaptability.

\red{Among all these controllers targeted at soft robotics mentioned above, the Jacobian controller is one of the simplest controllers for soft robotics \cite{14MY,24ZCa}.} 
Researchers were inspired by the Jacobian controller in rigid robot control and employed it in soft robots. 
Prior to actuation, the initial Jacobian matrix is required, and the authors in \cite{14MY} estimate it by actuating each actuator independently by an incremental amount.
Then, the actuation is determined by either optimization employing the Jacobian matrix as the forward model \cite{17KLb} or calculation using the inverse Jacobian matrix \cite{18ML}. 
Following each actuation step, the Jacobian matrix updates according to the previous robot states and actions, allowing it to adapt to local dynamics. Once the update finishes, the control loop proceeds with the next iteration.

Several modifications have been introduced to employ this controller in real-world applications and achieve better performance. 
For instance, robot states with a larger update interval are utilized in \cite{17MY} for the Jacobian matrix update. In this case, the disturbance can be rejected, and such a controller can be applied for complex cardiac ablation tasks. 
In \cite{17YJ}, the Jacobian update process is smoothed by averaging the previous Jacobian matrix with the newly estimated one, resulting in more stable motion control.
To perform segmented soft robot control, the authors in \cite{17TL} bridge the gap between robot end position and actuation via segment orientations instead of directly mapping between end position and actuation. In this way, complicated nonlinear mapping functions are decomposed to fit the Jacobian matrix. 

While the adaptations above have extended the Jacobian controller's capabilities to support complex tasks, they also reflect a fundamental limitation: soft robots often deviate from the original Jacobian model and its underlying assumptions. 
Specifically, the Jacobian approaches suppose that the robot state change $\triangle s$ is linear to the actuation change $\triangle a$, expressed as $\triangle s = J \triangle a$. 
However, considering the time delay in soft robot motion, the robot state may continue to change ($\triangle s \neq 0$) when the actuation remains unchanged ($\triangle a = 0$).
Due to the unsatisfactory performance caused by misalignment, the Jacobian matrix is primarily applied for motion description in principle, even with the adaptations mentioned above, and other data-driven controllers (NN in most cases) replace Jacobian controllers in practice. 
For instance, the Jacobian matrix is referenced in \cite{17TTb} for the theoretical soft robot modeling, but a neural network is applied for control to replace the Jacobian approach. 

Compared to straightforward yet limited Jacobian controllers, NNs have emerged as the most widely adopted approach in soft manipulator control \cite{24ZCa}.
Their suitability stems from the inherent nonlinearity and delay of soft robotic motion, which aligns well with the nonlinear activation functions and complex architectures of neural networks.
Moreover, the rich network structure and the large number of independent trainable parameters are suitable to cope with the intricate soft robot motion. 
Basic NNs like multilayer perceptron (MLP) can be applied to handle the delayed soft robot motion \cite{21GL}.
Furthermore, RNNs are composed of recurrent structures developed specifically for sequential problems, such as hysteresis \cite{18TT} and segment sequence \cite{23ZCd} in soft robot control. 

In this work, \red{based on the novel analysis of the compact Jacobian controller and accurate RNN controller}, we introduce an adaptive extended Jacobian controller, {\textit{AdapJ}}, specifically for soft manipulators. 
We retain the compact structure of the Jacobian controller while relaxing the parameter constraints. 
Also, we update the controller directly instead of the forward model and detail the matrix initialization and update approach learned from NN and model predictive controller (MPC). 
As illustrated in Fig. \ref{fig1.1}, AdapJ requires significantly less training data than most existing soft manipulator controllers owing to its concise architecture.
Despite its simplicity, AdapJ demonstrates superior performance in the simulation and real-world experiments, \red{outperforming classical and state-of-the-art (SOTA) controllers such as the Jacobian controller \cite{14MY}, MPC \cite{19ZT}, RNN-based controllers \cite{23ZCc}, and the iterative feedback controller (IFC) \cite{24XL}.}
In addition, its capacity for online updates endows it with adaptability to different frequencies, physical parameters, and even external disturbances.

The contributions of this paper are as follows:
\begin{enumerate}
\item 
\red{Based on the insightful analysis of the Jacobian controller and RNN controller, we propose an adaptive extended Jacobian controller for soft manipulators. Inspired by RNN, it relaxes the parameter constraints inherent in traditional Jacobian methods to better address hysteresis.}
\item 
{We leverage the inverse dynamics, rather than the corresponding forward model in the Jacobian controller, for initialization and online updating.
Inspired by RNN and MPC, we apply motor babbling and the Gauss-Newton algorithm for initialization and online updating.}
\item 
\red{We conduct simulation experiments to compare AdapJ with existing controllers such as the Jacobian controller, MPC, RNN-based controllers, and IFC.
Also, we analyse the controller parameter adaptation in response to variations in robot stiffness and damping properties.
Real-world experiments on a soft robot manipulator further demonstrate the superior performance of our controller over IFC, MPC, and RNN baselines.}
We also validate the controller’s adaptability under different frequencies, physical properties, and external disturbances.

\end{enumerate}

The rest of the paper is structured as follows: 
\red{Sec. \ref{sec2} introduces AdapJ, beginning with a novel analysis of the limitations and advantages of the Jacobian and RNN controllers. It also details the initialization and updating principles of AdapJ.}
Sec. \ref{sec3} describes the simulation and real-world experimental setup, including robots, working spaces, and devices. 
{Sec. \ref{sec4} compares our controller with some existing controllers and analyzes the controller parameter adaptation.
Then, we demonstrate} the effectiveness of our controller through real-world experiments. The adaptability is also validated across various scenarios. 
Sec. \ref{sec5} summarizes the work and outlines potential directions for future research.

\section{Controller Design}
\label{sec2}
In this section, we begin by analyzing the original Jacobian controller and RNN controller in Sec. \ref{sec2.1}. \red{Based on the insightful analysis}, we propose our adaptive extended Jacobian controller for soft manipulators in Sec. \ref{sec2.2}, followed by the proposed initialization and update strategy in Sec. \ref{sec2.3}.

\subsection{Existing Controller Analysis}
\label{sec2.1}
\subsubsection{Jacobian controller}
\label{sec2.1.1}
\begin{figure}[!ht]
\centering
\includegraphics[width=\linewidth]{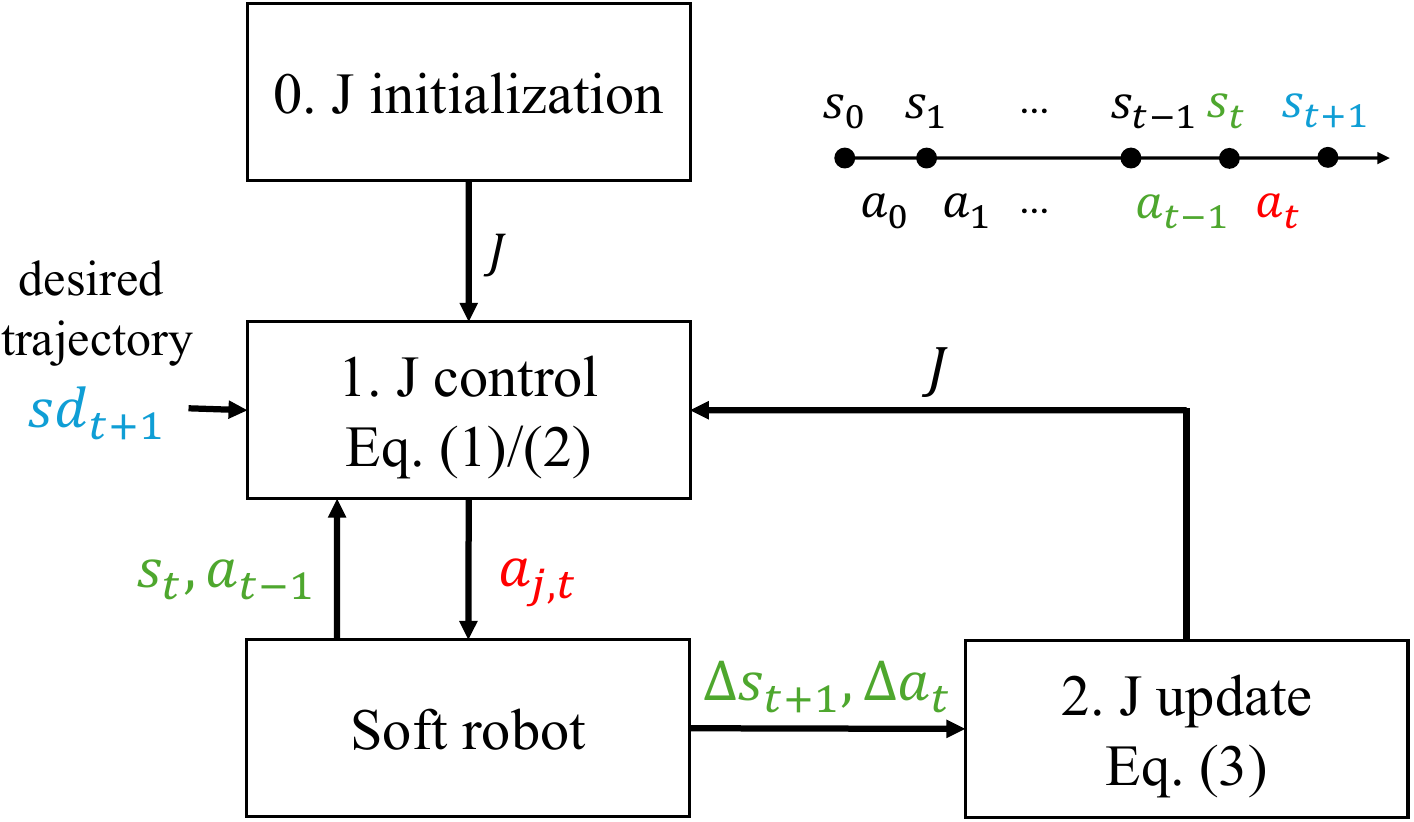}
\caption{Diagram of the standard Jacobian controller. Following initialization of the Jacobian matrix, the Jacobian controller computes the actuation $a_{j,t}$ (red) based on the previous robot state, actuation $s_t, a_{t-1}$ (green), and desired robot state $sd_{t+1}$ (blue). Then the Jacobian matrix updates according to the difference in robot state and actuation $\triangle s_{t+1}, \triangle a_t$ (green).}
\label{fig2.1}
\end{figure}

The Jacobian controller in \cite{14MY} is shown in Fig. \ref{fig2.1}. First, the Jacobian controller is initialized by moving each actuator independently by an incremental amount while measuring the motion.
In addition to this data collection method, some researchers also initialize the Jacobian matrix via a physical kinematics model \cite{17YJ}.
After initializing the Jacobian matrix $J\in R^{d_s\times d_a}$, where $d_s$ and $d_a$ denote the dimensions of state and actuation, the following optimal control strategy is implemented. Combining different strategies in \cite{14MY, 17KLb, 20YW}, the actuation ${a}_{j,t}$ is determined by
\begin{equation}
\label{eq2.1}
\begin{split}
\min_{{a}_{j,t}}\ &\alpha_1 \Vert {a}_{j,t}\Vert_2 + \alpha_2 \Vert \triangle{a}_{j,t}\Vert_2\\
s.t.\ &\triangle sd_{t+1} = J \triangle a_{j,t},\\
&\triangle sd_{t+1} = sd_{t+1} - p_{t},\\
&\triangle a_{j,t} = {a}_{j,t} - a_{t-1},\\
\end{split}
\end{equation}
where $s_t\in R^{d_s}, a_{t-1}\in R^{d_a}$ represent previous robot state and actuation at $t$-th and $t-1$-th timestep, and {the robot state $s$ is robot end position in most works \cite{14MY} and bending angle in some others \cite{19ZT}.}
$sd_{t+1}$ denotes the desired robot state from the desired trajectory. $\alpha_*$ represent cost weights, where $\alpha_1=1, \alpha_2=0$ in \cite{14MY} and $\alpha_1=0, \alpha_2=1$ in \cite{20YW}.

In addition to this optimal control strategy, the action can also be decided by using the inverse Jacobian matrix $J^{-1}\in R^{d_a\times d_s}$\cite{18ML}, which can be denoted as
\begin{equation}
\label{eq2.2}
\begin{split}
{a}_{j,t} &= J^{-1} (sd_{t+1} - s_{t}) + a_{t-1}\\
          &= J^{-1} sd_{t+1} + (-J^{-1}) s_{t} + I a_{t-1}.\\
\end{split}
\end{equation}

After the action decision, the Jacobian matrix will be updated according to the feedback. The strategy combined from \cite{17KLb, 14MY, 24CP} can be denoted as
\begin{equation}
\label{eq2.3}
\begin{split}
\min_{\triangle J}\ &\beta_1 \Vert \triangle s_{t} - (J + \triangle J) \triangle a_{t-1} \Vert_2 + \beta_2 \Vert \triangle J \Vert_2\\
s.t.\ &\triangle s_{t} = s_{t} - s_{t-1},\\
&\triangle a_{t-1} = a_{t-1} - a_{t-2},\\
\end{split}
\end{equation}
where $J + \triangle J$ represent the new Jacobian matrix. $\beta_*$ represent cost weights, where $\beta_1=0, \beta_2=1$ in \cite{14MY}, $\beta_1=1, \beta_2=0$ in \cite{17KLb}, and $\beta_1=\beta_2=1$ in \cite{24CP} for robust Jacobian estimation.

\begin{figure}[!ht]
\centering
\includegraphics[width=\linewidth]{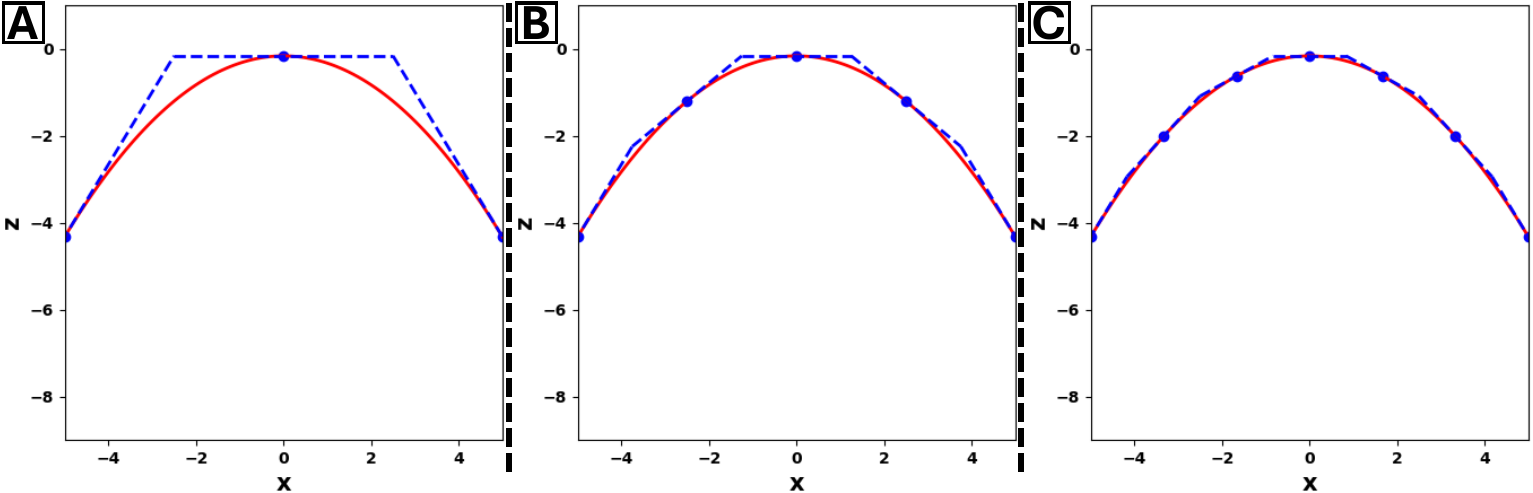}
\caption{
{The diagram of the Jacobian principle. 
The nonlinear function $z=-(x^2+1)/6$ is approximated by (A) 3, (B) 5, and (C) 7 linear functions.
The blue dotted lines represent the linear approximation functions, and the red lines represent the nonlinear function.}
}
\label{fig2.2}
\end{figure}

{In addition to describing the controller, we aim to illustrate the core principle of the Jacobian approach—approximating nonlinear functions using multiple local linearizations—in Fig. \ref{fig2.2}.
Specifically, we approximate the nonlinear function $z=-(x^2+1)/6$ with multiple linear functions at local sampling points. 
As the number of sampling points increases from Fig. \ref{fig2.2}-(A) to Fig. \ref{fig2.2}-(C), the Jacobian strategy achieves better approximation performance, aligning with the high control frequency (1 kHz) in the classical Jacobian controller work \cite{14MY}. 
}

\red{After introducing the Jacobian controller and its core principle, we analyze its shortcomings.}
First, the Jacobian assumption $\triangle s = J \triangle a$ does not suit soft robot motion because the robot may move ($\triangle s \neq 0$) when the actuation keeps constant ($\triangle a = 0$), considering the delay of soft manipulator motion.
In other words, the change in state is not solely determined by the actuation difference within a single timestep.
\red{In addition, there are strong constraints on the partial derivatives in Eq. (\ref{eq2.2}), which are} 
\begin{equation}
\label{eq2.4}
\begin{split}
&\frac{\partial {a}_{j,t}}{\partial a_{t-1}}=I,\\
&\frac{\partial {a}_{j,t}}{\partial sd_{t+1}}=-\frac{\partial {a}_{j,t}}{\partial s_{t}}=J^{-1}.\\
\end{split}
\end{equation}
\red{The parameter coupling may impair the approximation of soft robot behaviors.} 

\subsubsection{RNN controller}
\label{sec2.1.2}
We denote the RNN controller as 
\begin{equation}
\label{eq2.5}
\begin{split}
{a}_{RNN,t} = f_{RNN}(sd_{t+1},s_t\sim s_{t-n+2},a_{t-1}\sim a_{t-n}),
\end{split}
\end{equation}
where $n$ denotes the timestep of the recurrent layers. ${a}_{RNN,t}$ is the actuation determined by the RNN controller and produced by a fully connected layer following the recurrent layers. 
Unlike the Jacobian controller, the RNN parameters are not subject to strong structural coupling constraints imposed by the Jacobian formulation, as described in Eq. (\ref{eq2.4}).
\red{Based on the observed differences in parameter coupling and the corresponding impact on control performance, we hypothesize that parameter independence may also influence the effectiveness of soft robot controllers. To validate this, we propose a simple mathematical model.}

\begin{figure}[!ht]
\centering
\includegraphics[width=\linewidth]{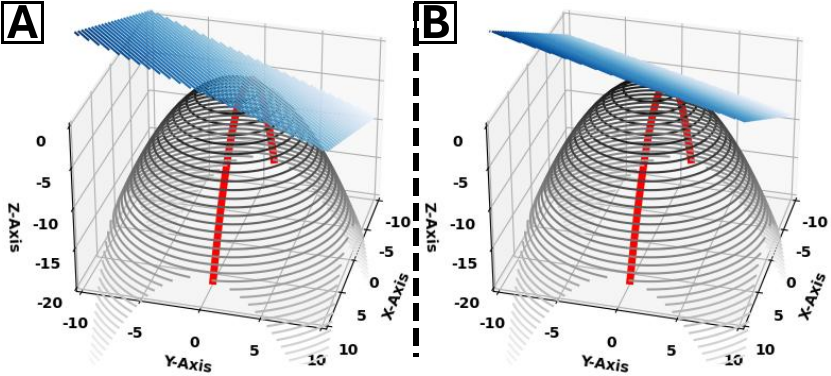}
\caption{{
The diagrams of approximating a multivariable nonlinear function using linear approximation with (A) coupling and \red{(B)} independent variables. 
The grey curved surfaces represent the multivariable nonlinear function $z=-(x^2+y^2)/6$, the red lines represent the nonlinear function $z=-(x^2+1)/6$, and the blue planes represent the linear approximation functions.}
}
\label{fig2.3}
\end{figure}

{Given the delayed response in soft robot motion, both modeling and control can be formulated as multivariable nonlinear functions that depend on states and actions over multiple timesteps.
We analyze how the parameter independence influences the approximation of multivariable nonlinear functions and furthermore address hysteresis, as shown in Fig. \ref{fig2.3}.
While the effect of sampling density (or control update frequency) on approximation accuracy has been discussed previously, here we focus on comparing approximation quality at a single point.
To illustrate this, we approximate the multivariable nonlinear function $z = -(x^2+y^2)/6$ at the point $(-2, 1, -\frac{5}{6})$ using linear functions with and without parameter coupling.
If we couple two variables by imposing the condition $\frac{\partial z}{\partial y}=-\frac{\partial z}{\partial x}$, analogous to the second Jacobian constraint in Eq. (\ref{eq2.4}), the linear approximation becomes $z = \frac{1}{2}x - \frac{1}{2}y+\frac{2}{3}$, shown in Fig. \ref{fig2.3}-(A).
If we do not impose any coupling between $\frac{\partial z}{\partial y}$ and $\frac{\partial z}{\partial x}$, the linear approximation is $z = \frac{2}{3}x - \frac{1}{3}y+\frac{5}{6}$, as illustrated in Fig. \ref{fig2.3}-(B).
}

\begin{table}[!ht]
\caption{{multivariable nonlinear function approximation errors (1e-3)}}
\centering
\begin{tabular}{l|l l}
variable&independent&coupling \\
\hline
error&$\bm{26.67}$&58.67\\
\end{tabular}
\label{table2.1}
\end{table}

{
To quantitatively evaluate approximation performance, we compute the deviation between the linear approximation and the true nonlinear surface at points sharing the same $x$ and $y$ coordinates. Specifically, we sample points with $x\in[-2.4, -2.2, -2, -1.8, -1.6],\ y\in[0.6,0.8,1,0.2,1.4]$, chosen for their proximity to the approximation point $(-2,1)$.
The average approximation errors for each linearization strategy are presented in Table \ref{table2.1}.
\red{The results indicate that the independent variable case outperforms the coupling cases, demonstrating that independent parameters improve the accuracy of linear approximations for multivariable nonlinear functions and soft robotics modeling and control.
Motivated by this insight, we propose the extended adaptive Jacobian controller, which builds upon the original Jacobian controller format but relaxes parameter constraints.}
}

\subsection{Adaptive Extended Jacobian Controller-AdapJ}
\label{sec2.2}

\begin{figure}[!ht]
\centering
\includegraphics[width=\linewidth]{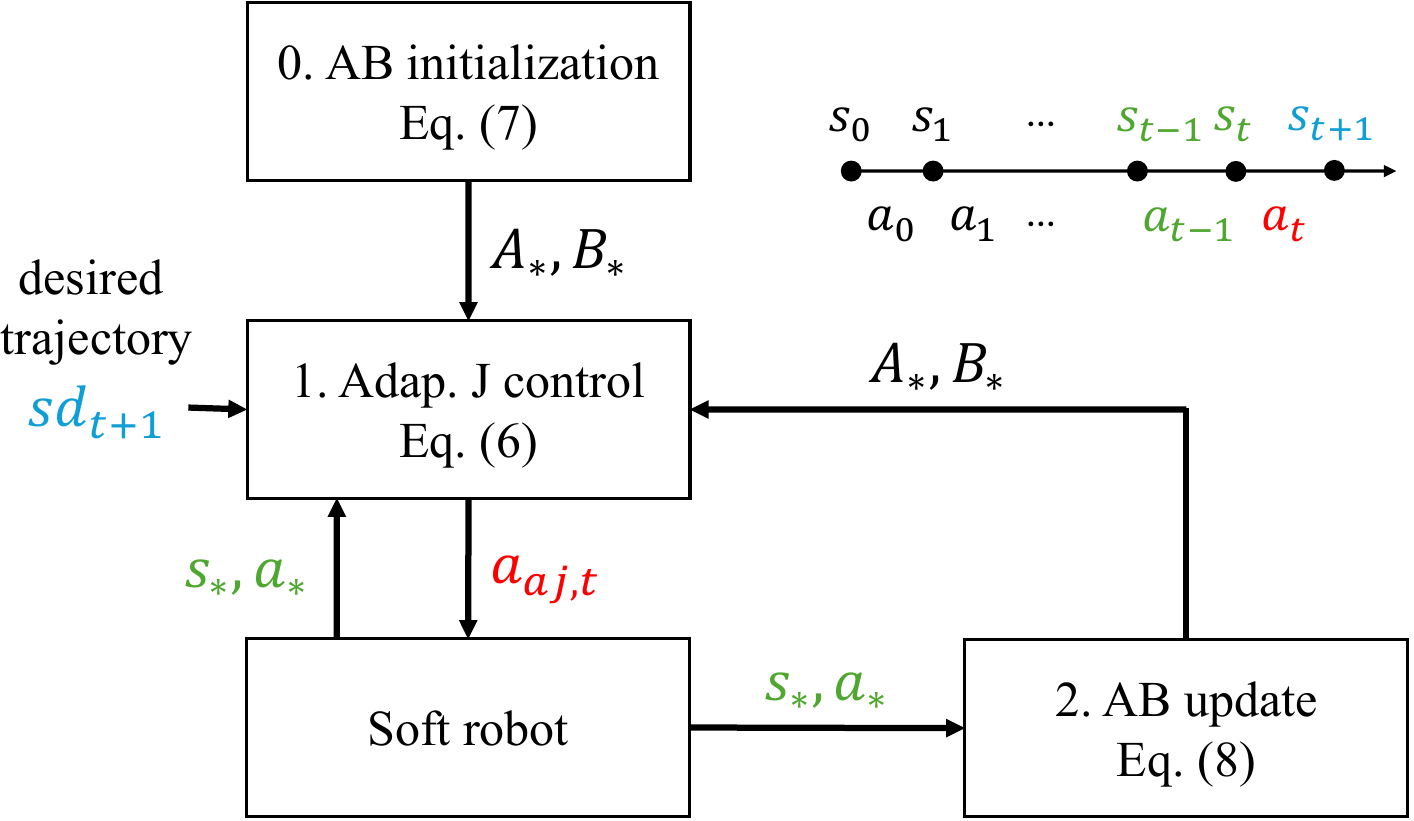}
\caption{
\red{Diagram of our adaptive extended Jacobian controller. The extended inverse Jacobian matrices $A_*, B_*$ are initialized with motor babbling and batch optimization. Then, the actuation $a_{aj,t}$ (red) is computed based on these matrices, previous robot states, actuations $s_*, a_*$ (green), and the desired robot state $sd_{t+1}$ (blue). After execution, the matrices $A_*, B_*$ are updated, and the control loop proceeds to the next iteration.}}
\label{fig2.4}
\end{figure}

Based on the insights from the existing controller discussion, we propose an adaptive extended Jacobian controller, AdapJ, as shown in Fig. \ref{fig2.4}. It can be denoted as 
\begin{equation}
\label{eq2.9}
\begin{split}
{a}_{aj,t} = A_0 sd_{t+1} + A_1 s_{t} + B_0 a_{t-1},
\end{split}
\end{equation}
where $A_*\in R^{d_a\times d_s}, B_*\in R^{d_a\times d_a}$ represent independent extended inverse Jacobian matrices and ${a}_{aj,t}$ represent the $t$-th timestep actuation decided by our controller. 
{Different from the conventional Jacobian controller, the matrices $\frac{\partial {a}_{aj,t}}{\partial sd_{t+1}}, \frac{\partial {a}_{aj,t}}{\partial s_{t}}, \frac{\partial {a}_{aj,t}}{\partial a_{t-1}}$ are independent and do not coupled following the Jacobian assumption in Eq. (\ref{eq2.4}).}
This adaptation is inspired by the independent parameters in RNN, and initially validated in the multivariable nonlinear function example in Fig. \ref{fig2.3}.
The proposed controller can be seen as an extended Jacobian controller because it will degrade to the inverse Jacobian controller in Eq. (\ref{eq2.2}) when $A_0=-A_1=J^{-1}, B_0=\bm{I}$.

\subsection{Matrix Initialization and Update}
\label{sec2.3}
{Considering the parameter independence, it is impossible to follow the initialization strategy in \cite{14MY} by actuating each actuator and directly calculating the matrices.
Therefore, we collect data from soft robots to initialize extended inverse Jacobian matrices $A_*, B_*$ by utilizing the motor babbling strategy like RNN controllers. }
Due to the proposed controller conciseness, it requires far fewer samples (only 100 in our cases) than RNN controllers (at least 5000), which means we collect $a_0\sim a_{99}$ and $s_0\sim s_{99}$ utilizing motor babbling strategy.
{The Jacobian control approach in \cite{14MY} initializes the forward Jacobian matrix. Similarly, MPC in \cite{19ZT} leverages a forward model and optimization for actuation determination.
However, we omit the forward model and directly initialize the inverse dynamics.
The extended inverse Jacobian matrices $A_*,B_*$ can be initialized as
\begin{equation}
\label{eq2.10}
\begin{split}
\min_{A_*,B_*}\ &\Vert {a}_{aj,t}-a_t\Vert\\
s.t.\ &{a}_{aj,t} = A_0 s_{t+1} + A_1 s_{t} + B_0 s_{t-1}\\
\end{split}
\end{equation}
}where $t\in[1, 98]$. 
For instance, when $t=1$, we estimate ${a}_{aj,1}$ based on the robot states and actuations $s_{2}, s_{1}, a_{0}$. When $t=98$, we estimate ${a}_{aj,98}$ based on the robot states and actuations $s_{99}, s_{98}, a_{97}$.
We learn from RNN and employ batch optimization to stabilize the initialization process.
By updating the matrices online, AdapJ can approximate the real inverse dynamics with the nominal one as an initial guess step by step, endowing it with adaptability to fixed physical property changes such as stiffness and damping changes.

{Following the MPC updating principle in \cite{19ZT}, the Gauss-Newton method is applied for a short computational time, and the updating strategy can be denoted as}
{
\begin{equation}
\label{eq2.11}
\begin{split}
\omega = 
\begin{cases}
\omega + \triangle \omega, & \Vert \triangle \omega \Vert 
\leq \red{\triangle \omega^{\max}},\\
\omega + \frac{\triangle \omega}{\Vert \triangle \omega \Vert} \red{\triangle \omega^{\max}},   & \Vert \triangle \omega \Vert>\red{\triangle \omega^{\max}},\\
\end{cases}\\
\end{split}
\end{equation}
}
{where}
{
\begin{equation}
\label{eq2.12}
\begin{split}
\triangle \omega &= \rho {(\phi\phi^T)}^{-1}\phi E,\\
\omega &=[A_0\ A_1\ B_0]^T\in R^{(2d_s+d_a)\times d_a}, \\
\phi &=[s_{t+1}\ s_{t}\ a_{t-1}]^T\in R^{(2d_s+d_a)\times1}, \\
E &=[a_{t} - {a}_{aj,t}]^T\in R^{1\times d_a},\\
\end{split}
\end{equation}
}
{and $\red{\triangle \omega^{\max}}$ denotes the maximum parameter change allowed at each update step for smooth update.
The constraint endows AdapJ with robustness to dynamic disturbance, which only affects the manipulator in several timesteps.
After $a_t$ is executed and $s_{t+1}$ is measured, the controller can be seen as an online actuation estimator, and we can update matrices $A_*, B_*$ following Eq. (\ref{eq2.11}) to fit the current local situation. The model parameter convergence utilizing the Gauss-Newton method for updating has been proven in \cite{19ZT}.}

\red{In our controller design, nonlinearity is addressed through online updates, as illustrated in Fig. \ref{fig2.2}, while hysteresis is mitigated by including previous state and actuation as input and relaxing parameter coupling constraints, as shown in Fig. \ref{fig2.3}.}
Although AdapJ is fundamentally a linear controller, it effectively approximates the nonlinear and delayed dynamics of soft manipulator behavior as validated in the following experiments.
Compared to Jacobian controllers, AdapJ removes parameter coupling to improve multivariable nonlinear function approximation.
Compared to RNN controllers, AdapJ employs significantly fewer parameters, enabling online updates and adaptability to fixed physical property changes and dynamic disturbances.
Compared to MPC, AdapJ directly approximates the inverse dynamics instead of the forward modeling, eliminating the necessity for actuation decision through time-consuming optimization in \cite{19ZT}.
Collectively, these design improvements result in an adaptive and precise controller well-suited for soft manipulator applications.

\section{Experiment Setup}
\label{sec3}
In this section, we introduce the experiment setup, devices, and the corresponding working space in simulation and the real world in Sec. \ref{sec3.1} and Sec. \ref{sec3.2}, respectively.

\subsection{Simulation Experimental Setup}
\label{sec3.1}

\begin{figure}[!ht]
\centering
\includegraphics[width=\linewidth]{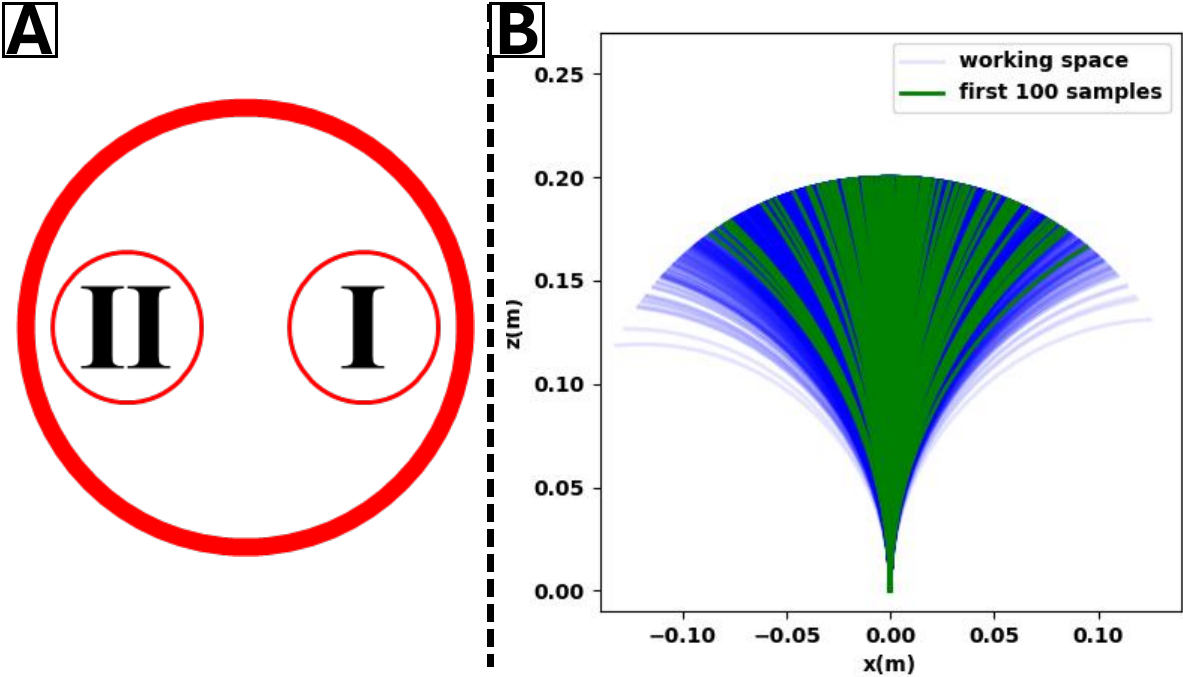}
\caption{
(A) Two actuation units are placed symmetrically in the robot.
(B) The working space in the simulation. 
The first 100 samples (green) are applied for the initialization of the matrices $A_*, B_*$.
The 5000 samples (light blue) collected by the motor babbling strategy basically show the robot working space and are applied to the other controller initialization.
}
\label{fig3.1}
\end{figure}

In simulation, we aim to compare controllers preliminarily and perform an intuitive parameter analysis.
Therefore, we built a simple soft robot finger based on the pseudo-rigid model in \cite{18CS}, whose length is 200 mm. The bending stiffness is 0.4 $Nm^2$ and the damping coefficient is 1 $Nms$.
As shown in Fig. \ref{fig3.1}-(A), two actuation units are placed symmetrically in the robot and actuated by $a_h\in R$,
where
\begin{equation}
\label{eq3.1}
\begin{split}
a_I &= \max\{0, a_h\}, \\
a_{II} &= \max\{0, -a_h\}.
\end{split}
\end{equation}
We randomly actuate the robot and collect 5000 samples, shown in Fig. \ref{fig3.1}-(B), and only the first 100 samples are applied for our controller initialization.
\red{The x position of the finger end is applied as the robot state $s\in R$, whose range is [-131.9, 131.9] mm.}
The actuation $a_h$ and the state $s$ are rescaled to [-1,1] for RNN.

\subsection{Real Experimental Setup}
\label{sec3.2}

\begin{figure}[!ht]
\centering
\includegraphics[width=\linewidth]{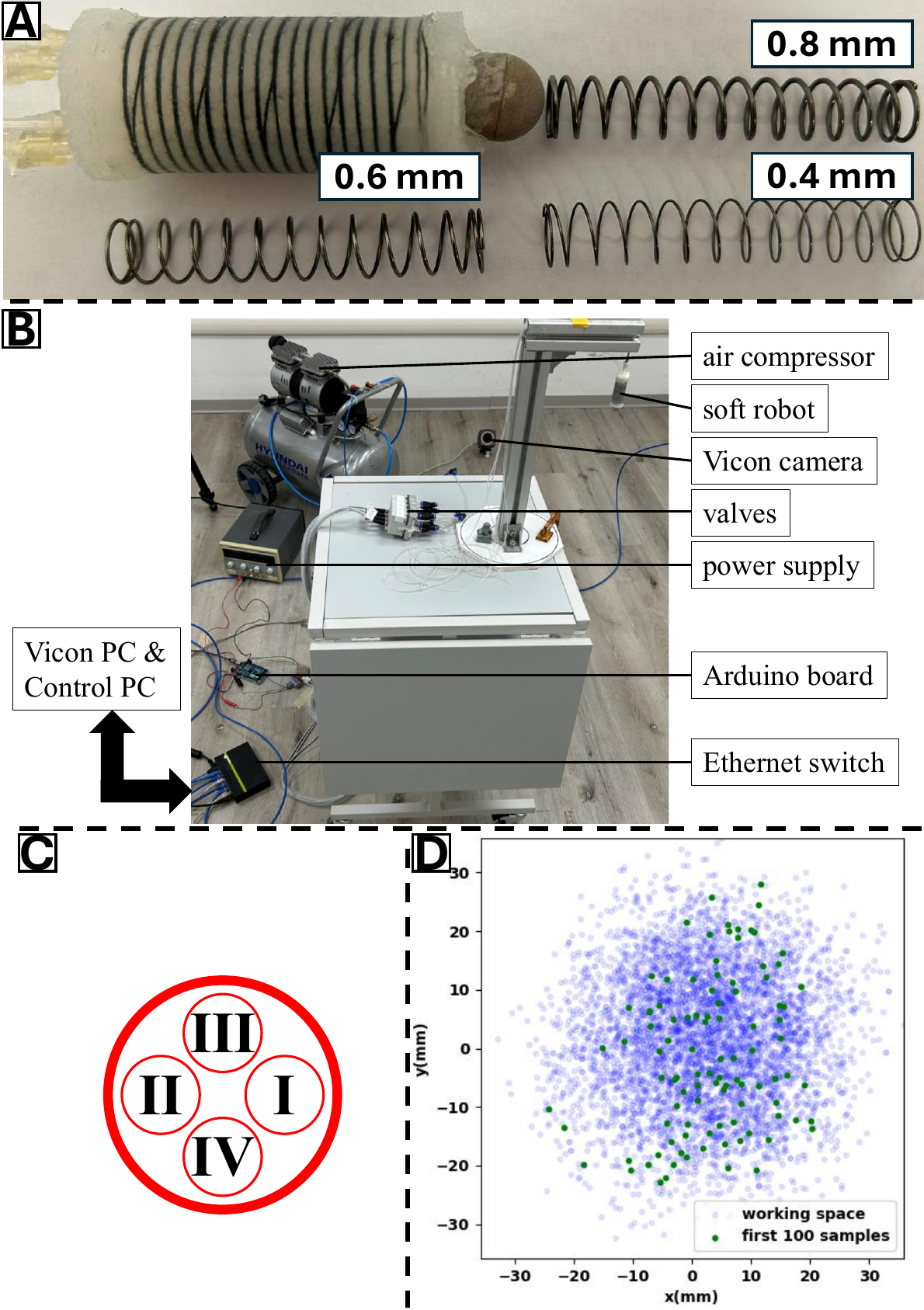}
\caption{
(A) Pneumatic soft manipulator in real experiments. One manipulator is composed of four chambers, and one springer is inserted inside the center for stiffness adjustment. Three kinds of springs (wire diameter 0.4mm, 0.6mm, and 0.8mm) are applied in our experiments.
(B) The experimental setup. The manipulator is actuated by an air compressor and controlled by valves and an Arduino board. The optical tracking system collects the motion.
(C) Robot actuation diagram. The actuation values in four actuation units $a_I, a_{II}, a_{III}, a_{IV}$ are controlled by the actuation $a=[a_h, a_v]$.
{(D) The working space in the real experiment. The first 100 samples (green) are applied for $A_*, B_*$ initialization, and 5000 samples (blue) are collected for the others. }
}
\label{fig3.2}
\end{figure}

To validate our controller in the real world, we manufacture a pneumatic silicone soft manipulator shown in Fig. \ref{fig3.2}-(A).
The robot is made of Ecoflex 00-30  (Smooth-On, Macungie, PA), and the length of each module is about 45mm.
There are four symmetrical chambers placed at an interval angle of $90^\circ$ along the circumferential direction inside the manipulator as shown in Fig. \ref{fig3.2}-(C). 
The cotton line wraps the robot to constrain the radial expansion. 
Springs (wire diameter 0.6 mm) are put inside the robot center to adjust the stiffness, and we name it Robot 1.  
The robot structure design can be found in \cite{24XR}.
To validate the adaptability of our controller, we change the inner spring (named Robot 2 for the wire diameter of 0.4 mm and Robot 3 for the wire diameter of 0.8 mm) to change the robot stiffness.
{The state $s\in R^2$ is the robot end position} and actuation $a=[a_h, a_v]\in R^2$ actuates four chambers following
\begin{equation}
\label{eq3.2}
\begin{split}
a_I &= \max\{0, a_h\},\\
a_{II} &= \max\{0, -a_h\},\\
a_{III} &= \max\{0, a_v\},\\
a_{IV} &= \max\{0, -a_v\}.\\
\end{split}
\end{equation}
Similar to simulation, they are rescaled to $[-1,1]$ for RNN.

The manipulator is actuated in the setup shown in Fig. \ref{fig3.2}-(B). An optical tracker ball is fixed at the end of the robot. Six optical tracking cameras (VICON Bonita) and one data collection computer named Vicon PC are used to track the robot motion. An air compressor connects chambers, whose pressures are controlled by valves (Camozzi K8P-0-E522-0). We control chamber pressure via a control PC (Ubuntu 20.04, CPU i5-12500H, and RTX 3050) and an Arduino MEGA board. Two computers and six cameras communicate through an Ethernet switch. Pytorch is applied for optimization and NN training. The data collection and control frequency is 10 Hz without a special statement.
Similar to the simulation, we collect 5000 samples to train the RNN controller and show the working space of Robot 1 in Fig. \ref{fig3.2}-(D). 
{The working space area is about $71.69mm\times 71.69mm$.}
The first 100 samples are employed for $A_*, B_*$ matrices initialization.

\section{Experiment Results}
\label{sec4}
In this section, we first compare \red{the original Jacobian controller \cite{14MY}, MPC \cite{19ZT}, RNN \cite{23ZCc}, IFC \cite{24XL},} and the proposed controller in simulation in Sec. \ref{sec4.1}.
The parameter adaptation in response to the robot's physical properties is also discussed.
Then we compare the proposed controller, MPC, \red{IFC,} and the RNN controller in the real-world experiments in Sec. \ref{sec4.2}. 
We validate the adaptability of our proposed controller in different frequencies, physical properties, and external disturbances.

\subsection{Preliminary Controller Analysis in Simulation}
\label{sec4.1}

In this section, we aim to compare the original Jacobian controller in Eq. (\ref{eq2.2}), the MPC in \cite{19ZT}, the RNN controller in Eq. (\ref{eq2.5}) and \cite{23ZCc}, \red{IFC in \cite{24XL}}, and our controller in Eq. (\ref{eq2.9}).
\red{IFC is based on RNN and leverages the discrepancy between the target and actual states from previous steps for online compensation.}
In addition, we train one RNN with only the first 100 samples and name this controller 'RNN\_100' to validate the RNN data requirement.
Then, we test them in various scenarios and analyse the parameter change of our controller under different stiffness and damping factors.
The experiments are carried out in the simulation environment mentioned in Sec. \ref{sec3.1}.

\subsubsection{Controller Comparison}
\label{sec4.1.1}
\begin{figure*}[!ht]
\centering
\includegraphics[width=\linewidth]{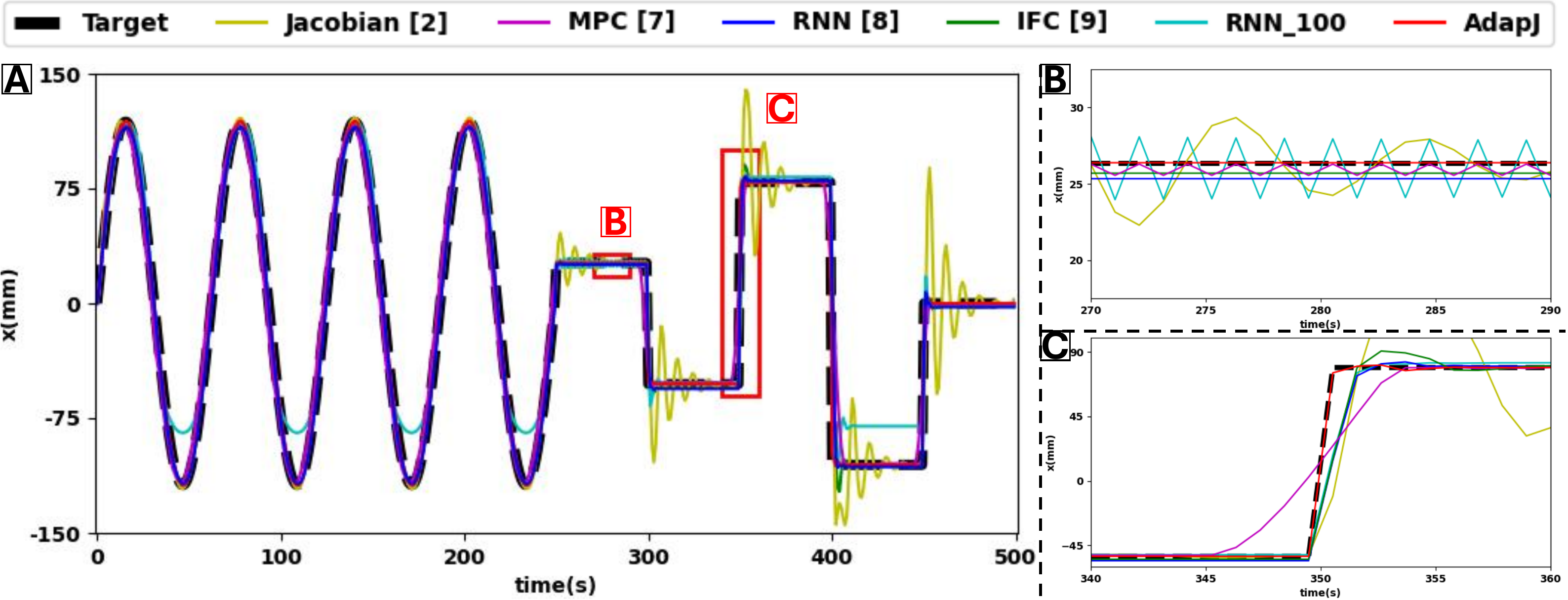}
\caption{
\red{(A) The target trajectory (dotted black) and the robot motion under the control of the original Jacobian controller \cite{14MY} (yellow), MPC \cite{19ZT} (magenta), RNN \cite{23ZCc} (blue), IFC \cite{24XL} (green), RNN\_100 (cyan), and our controller (red).
(B) and (C) show zoomed-in views of the regions highlighted by red boxes in (A), illustrating detailed behavior on (B) maintaining a static state at 26 mm and (C) abrupt target transition from -52 mm to 78 mm, respectively.
}
}
\label{fig4.1}
\end{figure*}

\red{First, we evaluate the controllers on a task involving sinusoidal tracking followed by stabilization at selected target angles, as shown in Fig. \ref{fig4.1}-(A). 
This task includes dynamic following, abrupt transition, and steady-state maintenance.} 
The tracking errors, computational times per step, number of parameters, and training times are reported in Table \ref{table4.1}. 
\red{All the errors included in this work are mean average errors.}
Compared to RNN, RNN\_100 exhibits reduced accuracy in both sine wave tracking and -105 mm maintenance tasks, as shown in Fig. \ref{fig4.1}-(A), demonstrating the necessity of a large amount of data for RNN.
\red{However, with the same amount of training data (100), AdapJ obviously outperforms any other controllers on all the various tasks, leveraging a concise linear format and online updating.}
The initial parameters are $A_0 = 2.89, A_1 = -1.86, B_0 = -0.55.$
This result indicates that soft robot motions do not follow the Jacobian principle, which is $A_0 = -A_1 =J^{-1}, B_0 = 1$.

\begin{table}[!ht]
\caption{\red{average errors, standard deviations, computational times per step, parameter numbers, and training times in simulation}}
\centering
\begin{tabular}{p{40pt}|p{45pt} p{40pt} p{30pt} p{35pt}}
&Error (mm)&Computational Time (ms)&{Parameter Number}&Traning Time (ms)\\
\hline
Jacobian\cite{14MY} &7.31$\pm$13.31     & $\bm{0.01}$ & $\bm{1}$ & $\bm{1.36}$\\
MPC\cite{19ZT}      &7.48$\pm$11.02     & 1.89        & 4        & 2.62\\
RNN\cite{23ZCc}     &2.34$\pm$6.81      & 0.25        & 13089    & 219.17\\
IFC\cite{24XL}      &1.99$\pm$6.87      & 0.26        & 13090    & 219.17\\
RNN\_100            &8.48$\pm$10.84     & 0.25        & 13089    & 78.29\\
Ours                &$\bm{0.30\pm0.56}$ & 0.04        & 3        & 2.02\\
\end{tabular}
\label{table4.1}
\end{table}

\red{Fig. \ref{fig4.1}-(B) and (C) illustrate the controller performance under steady-state maintenance and abrupt target transition with large deformation, respectively.
As shown in Fig. \ref{fig4.1}-(B), the Jacobian controller oscillates while gradually approaching the steady-state target, whereas MPC shares a similar behavior with reduced error.
Both RNN and IFC stabilize around 26 mm, with IFC achieving a lower error due to its online error compensation mechanism.
However, RNN\_100 exhibits persistent oscillations, indicating that RNN fails to capture inverse kinematics with limited data.
The degraded performance of RNN\_100 underscores a key limitation of offline data-driven approaches: their strong dependence on the quantity and coverage of training data, and insufficient data results in a significant loss of control accuracy.
Meanwhile, our AdapJ successfully accomplishes the steady-state maintenance task with the lowest error.}

\red{
In the abrupt target transition shown by Fig. \ref{fig4.1}-(C), the Jacobian controller still exhibits severe oscillations during the transition phase, and hence, a high deviation.
Under MPC control, the robot leaves the original target (-52 mm) before the target transition due to the predictive nature of MPC.
MPC, RNN, RNN\_100, and IFC reach the new target (78 mm) with noticeable delay, exhibiting overdamped responses and the soft robotics hysteresis.
Meanwhile, our proposed controller, trained with the same limited dataset, closely follows the target with minimal overshoot and fast convergence, demonstrating its superior convergence speed, high tracking precision, and strong data efficiency.}

\red{Considering both online computational cost and initialization time shown in Table \ref{table4.1}, the Jacobian approach is the fastest due to its simple structure, proving that it is one of the simplest controllers in soft robotics.
In the Jacobian controller, only a single matrix $J$ is required to be initialized and updated, whereas our controller updates three matrices $A_*, B_*$ sharing comparable size. 
Although RNN, RNN\_100, and IFC contain significantly more parameters than MPC and require longer initialization times ($>$70ms), they do not perform online updates.
In contrast, the model update and actuation decision via optimization in MPC \cite{19ZT} results in a longer time than direct actuation decision using RNN in Eq. \ref{eq2.5}.
Meanwhile, our controller achieves a shorter online computational time than all other controllers except the Jacobian controller, aligning with the statement in Fig. \ref{fig1.1}, and requires a short initialization time similar to MPC ($<$3ms).
}
\rred{The computational time is the average obtained from the entire motion simulation process, which lasts 500 seconds in the simulation environments and is performed on the PC described above.}

\subsubsection{Adaptability Validation}
\label{sec4.1.2}
\begin{figure}[!ht]
\centering
\includegraphics[width=\linewidth]{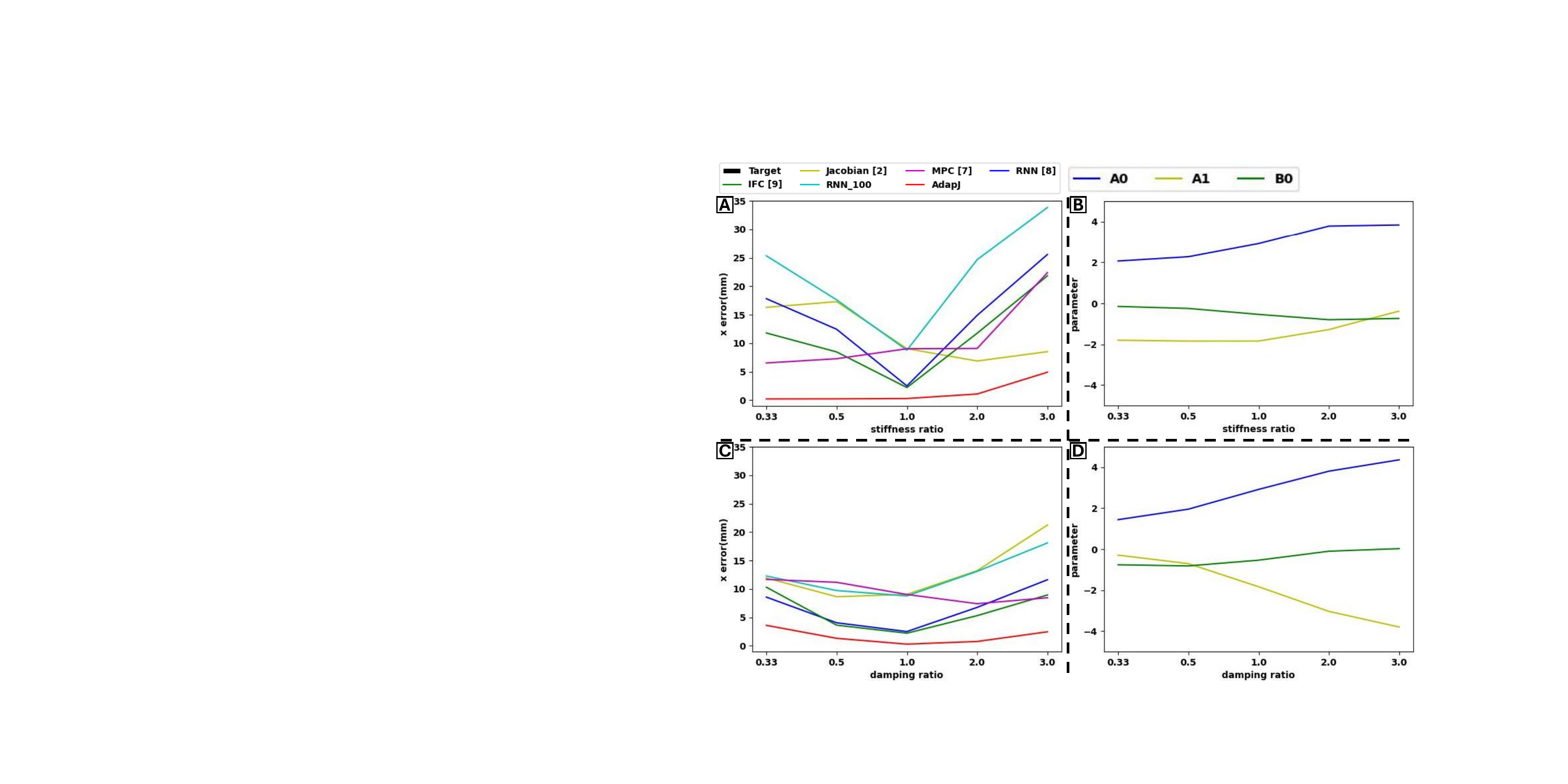}
\caption{
{The robot motion error under (A) stiffness and (C) damping change. The $A_*,B_*$ adjustment in response to the (B) stiffness and (D) damping change. The ratio 1 represents the original scenario, and a bigger ratio represents a higher stiffness/ damping coefficient.}
}
\label{fig4.2}
\end{figure}

After controller comparison in the default situation, we evaluate controller adaptability under different scenarios. Specifically, we modify the stiffness and damping based on the stiffness and damping ratios, and the average errors are depicted in Fig. \ref{fig4.2}-(A) and (C). 
The ratio 1 represents the default physical parameter, and a higher ratio represents a higher parameter.
RNN and RNN\_100, as offline controllers, induce higher tracking errors under different physical properties, especially with the stiffness change. The online updating Jacobian controller fails under some extreme situations, reflecting its limited adaptability. 
\red{Leveraging the online error compensation component, IFC outperforms the original RNN in most adaptability tests, and MPC also exhibits some adaptability to a certain extent, as indicated by the robust errors.
However, our controller still obviously outperforms the other controllers with low errors in all cases ($<$5mm).}

{We further analyze the adaptation of the parameters $A_*, B_*$ in response to the physical property change, as shown in Fig. \ref{fig4.2}-(B),(D). A simple soft robot finger is applied because its one-dimensional parameter space enables direct visualization.
The parameter $A_0$ corresponds to the feedforward term, and higher actuation is required in systems with higher stiffness or damping. Consequently, $A_0$ roughly increases as stiffness and damping coefficients rise.
For the discussion of $A_1$, we consider the simplified control expression $a_{aj,t} = A_0 0 + A_1s_t + B_00$, where the robot is expected to return to the original state $0$ from the current state $s_t$. A high-stiffness system tends to return to the equilibrium (0), hence $\Vert A_1\Vert$ decreases. 
Meanwhile, a high-damping system tends to resist motion and maintain the current state ($s_t$), and $\Vert A_1\Vert$ should be larger.
Regarding $B_0$, we consider the case $a_{aj,t} = A_0 0 + A_1 0 + B_0a_{t-1}$, and the feedforward term $B_0a_{t-1}$ serves as hysteresis compensation. In high-damping systems, motion is heavily constrained, leading to a lower $\Vert B_0\Vert$. In contrast, high-stiffness systems store more energy and exhibit stronger memory effects, necessitating a larger $\Vert B_0\Vert$ for effective compensation.}

\subsection{Real Experiment Results}
\label{sec4.2}

\subsubsection{Controller Comparison}
\label{sec4.2.1}
Using the real experiment setup in Sec. \ref{sec3.2}, we compare the control accuracy of \red{MPC, RNN, IFC, and AdapJ.} 
We first initialize the controller matrices $A_*, B_*$ following Eq. (\ref{eq2.10}) with only 100 samples and the other controllers with 5000 samples, as illustrated in Fig. \ref{fig3.2}-(D).
The initial extended inverse Jacobian matrix $A_*, B_*$ are
\begin{equation}
\label{eq4.1}
\begin{split}
&A_0=\left[\begin{matrix}
3.01&0.06\\
-0.03&3.00\\
\end{matrix}\right],
\ A_1=\left[\begin{matrix}
-1.79&-0.08\\
0.01&-1.77\\
\end{matrix}\right],\\
&B_0=\left[\begin{matrix}
-0.49&0.00\\
0.02&-0.57\\
\end{matrix}\right],\\
\end{split}
\end{equation}
which are much different from the Jacobian controller constraints, which are $A_0=-A_1=J^{-1}, B_0=\bm{I}$ mentioned in Eq. (\ref{eq2.4}).
Such a misalignment demonstrates that the original Jacobian controller is not suitable for soft robots.

\begin{figure*}[!ht]
\centering
\includegraphics[width=\linewidth]{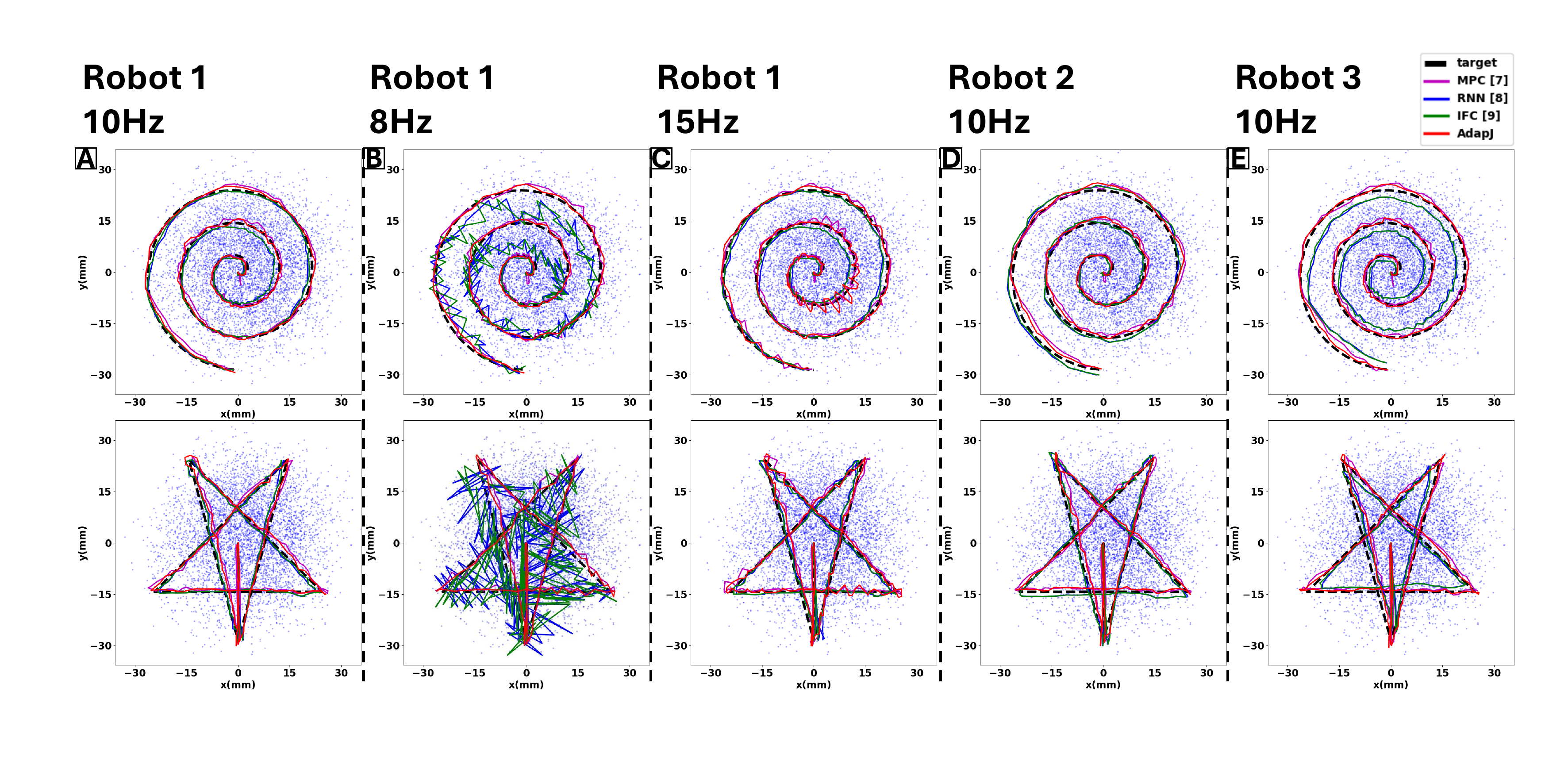}
\caption{
\red{The target trajectory (black) and measured robot motion of the (A) Robot 1 under 10Hz, (B) Robot 1 under 8Hz, (C) Robot 1 under 15Hz, (D) Robot 2, and (E) Robot 3 under the control of MPC \cite{19ZT} (purple), RNN \cite{23ZCc} (blue), IFC \cite{24XL} (green), and AdapJ (red).
The light blue dots show the robot's working space.
\red{4 (controller number) $\times$ 2 (trajectory number) $\times$ 5 (scenario number) $\times$ 3 (trial number for each condition) = 120 experiments are included.}}}
\label{fig4.3}
\end{figure*}

We evaluate the controllers on trajectory-following tasks using both a spiral and a \red{star-shaped} path, as shown in Fig. \ref{fig4.3}.
These trajectories are chosen to assess controller performance across curved segments, straight lines, and sharp turns. 
The robot motions are shown in Fig. \ref{fig4.3}-(A), and each experiment is repeated for three trials, as in the following experiments.
The average tracking errors, standard deviations, and the computational time per step are included in Table \ref{table4.2}. 
\red{Our controller is linear in structure and employs only 100 samples for initialization, far fewer than the samples used for RNN training (5000). Despite this, AdapJ achieves lower tracking errors than MPC, RNN, and IFC.
Meanwhile, AdapJ achieves the shortest computational time. The lower errors and shorter computational times are consistent with the simulation results and structure simplicity statement in Fig. \ref{fig1.1}.}
The computation times in the real-world experiments are longer than in the simulation due to the increased dimensionality of both the state and actuation spaces.

\begin{table*}[!ht]
\caption{\red{computational time per step (ms), average errors and standard deviations (mm) in the real experiments}}
\centering
\begin{tabular}{l|l l l l}
&MPC\cite{19ZT}&RNN\cite{23ZCc}&IFC\cite{24XL}&AdapJ\\
\hline
computational time &7.03&1.01&1.15&$\bm{0.35}$\\
\hline
spiral-10Hz     &1.16$\pm$0.78&1.26$\pm$0.53&1.13$\pm$0.48&$\bm{0.79\pm0.46}$\\
triangle-10Hz   &1.38$\pm$0.68&1.16$\pm$0.61&1.06$\pm$0.54&$\bm{0.92\pm0.53}$\\
\hline
spiral-8Hz      &1.57$\pm$0.90&5.24$\pm$4.56&5.27$\pm$4.39&$\bm{0.89\pm0.52}$\\
triangle-8Hz    &1.80$\pm$0.95&6.65$\pm$4.21&6.43$\pm$3.95&$\bm{1.09\pm0.68}$\\
\hline
spiral-15Hz     &1.28$\pm$0.85&1.75$\pm$0.64&1.49$\pm$0.56&$\bm{1.23\pm0.89}$\\
triangle-15Hz   &1.27$\pm$0.71&1.76$\pm$0.74&1.43$\pm$0.62&$\bm{1.22\pm0.82}$\\
\hline
spiral-Robot 2  &1.41$\pm$0.86&1.34$\pm$0.64&1.17$\pm$0.57&$\bm{0.97\pm0.61}$\\
triangle-Robot 2&1.26$\pm$0.65&1.31$\pm$0.58&1.11$\pm$0.53&$\bm{0.93\pm0.53}$\\
\hline
spiral-Robot 3  &1.31$\pm$0.89&2.39$\pm$0.85&2.04$\pm$0.75&$\bm{0.81\pm0.49}$\\
triangle-Robot 3&1.43$\pm$0.76&2.39$\pm$1.04&2.01$\pm$0.90&$\bm{0.98\pm0.60}$\\
\end{tabular}
\label{table4.2}
\end{table*}

\subsubsection{Adaptability Validation}
\label{sec4.2.2}

{The simple structure of our controller makes it possible to update the matrices $A_*, B_*$ online and adjust to situations different from the initial one, which cannot be achieved by offline controllers like RNN. 
In this case, we include different frequencies (8 Hz and 15 Hz) and different stiffnesses (Robots 2 and 3) to validate the adaptability of our controller.
The robot motions under the control of MPC, RNN, IFC, and the proposed controller are shown in Fig. \ref{fig4.3}-(B)$\sim$(E). The errors and standard deviations are included in Table \ref{table4.2}.}

Frequency changes induce misalignment between the training and the test situation. With limited inference ability and a lack of online updating, the RNN controller can roughly follow the trajectories but produces aggressive motions and higher errors, as shown in Fig. \ref{fig4.3}-(B) and (C). 
IFC, composed of RNN and an online compensation mechanism, achieves adaptability with lower errors than the original RNN.
\red{Meanwhile, MPC and our controller can update online and follow the trajectories with different frequencies and lower errors, while AdapJ achieves the lowest tracking errors in these controllers.}

Robot 2 is softer than Robot 1, and Robot 3 is harder than Robot 1. 
Therefore, in the trajectory following tasks, the RNN controller trained on Robot 1 will produce larger real trajectories on Robot 2 and smaller real trajectories on Robot 3, as shown in Fig. \ref{fig4.3}-(D) and (E). 
\red{Similarly, AdapJ achieves better tracking accuracy than the other controllers in the physical property adaptability experiments.
Overall, under these diverse scenarios (different frequencies, different stiffness, and different damping in simulation), our controller can update online and follow trajectories with the lowest errors, demonstrating its adaptability.}

\subsubsection{Disturbance and Obstacle Rejection}
\label{sec4.2.3}

\begin{figure}[!ht]
\centering
\includegraphics[width=\linewidth]{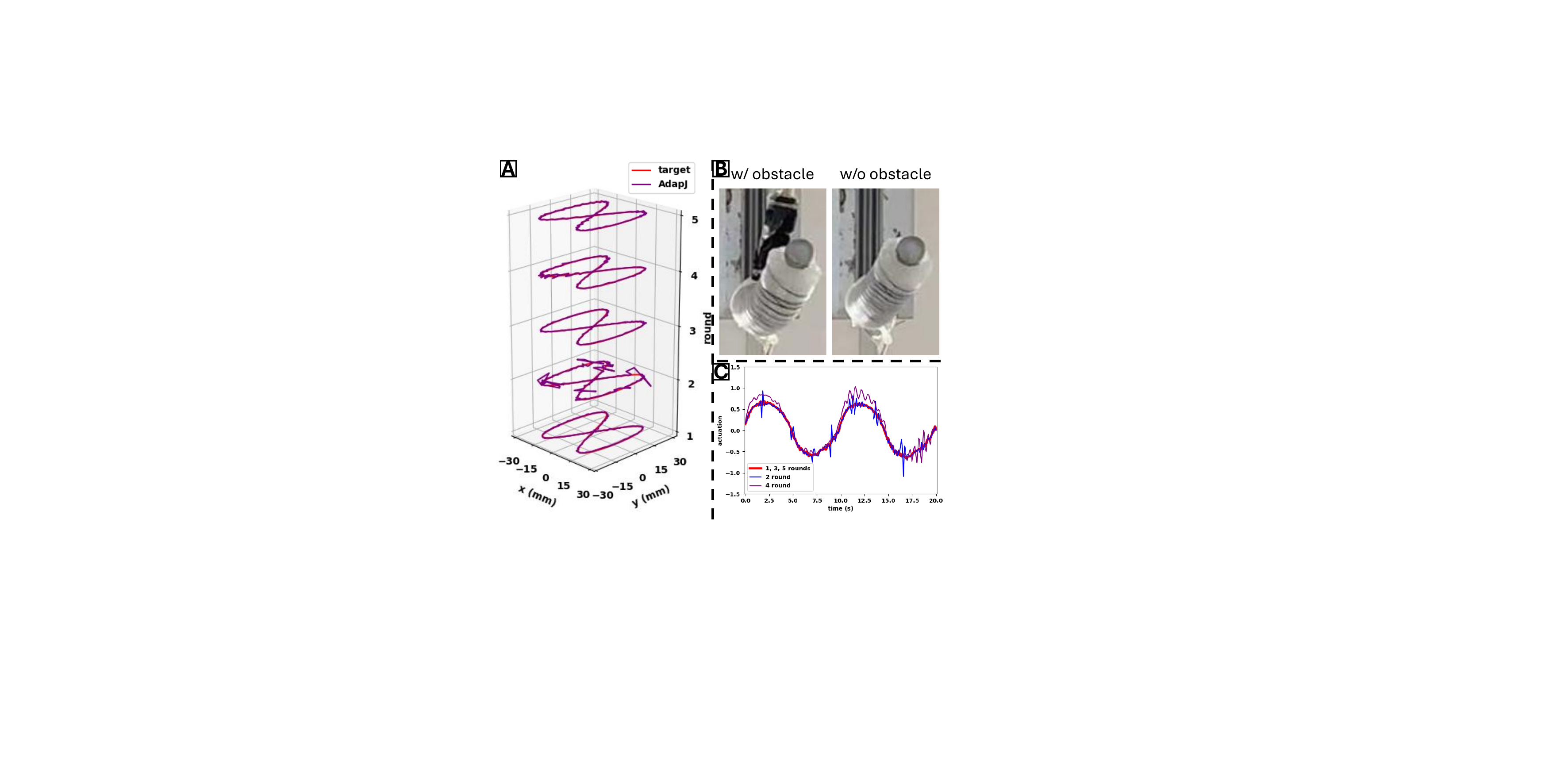}
\caption{
{(A) The target (red) and measured (purple) trajectories under the control of AdapJ in five rounds.
(B) The soft manipulator deformation with and without an obstacle.
(C) The average $a_v$ in the 1st, 3rd, and 5th rounds without obstacle (red) and $a_v$ in the 2nd round under dynamic disturbance (blue) and in the 4th round with an obstacle (purple). }
}
\label{fig4.4}
\end{figure}

{In addition to the frequency and physical property adaptability, we also conduct experiments to validate the robustness of our controller under dynamic disturbance and fixed obstacles.
In this experiment, the manipulator is controlled to follow a Lissajous trajectory for five rounds, as shown in Fig. \ref{fig4.4}-(A).
In the first round, the manipulator can follow the trajectory.
In the second round, we continuously and randomly pitch the manipulator.
Under the dynamic disturbance, the manipulator can still roughly follow the trajectory and recover the following accuracy in the third round without disturbance.
In the fourth round, we fix an obstacle near chamber III, as shown in Fig. \ref{fig4.4}-(B).
In this case, AdapJ follows the trajectory by online updating and decides on a larger $a_v$, as illustrated in Fig. \ref{fig4.4}-(C).
After the adaptation to the obstacle, we remove it in the fifth round, and AdapJ can still follow the target trajectory.
The experiment results in Fig. \ref{fig4.4} demonstrate that the AdapJ is adaptive to dynamic disturbances and fixed obstacles with the help of the extended Jacobian format and the online updating strategy.
The experiment video is included in the Supplemental Video.}

\section{Conclusion and Discusssion}
\label{sec5}
In this work, we return to one of the simplest soft manipulator controllers, the Jacobian controller, and propose an extended Jacobian controller adaptive to different situations. 
Such a controller is inspired by the linear format of the Jacobian controller and the parameter independence of the RNN controller.
We also employ motor babbling and batch optimization as the controller initialization strategy following the RNN controller and the Gauss–Newton method as the updating strategy following MPC. 
As mentioned in Fig. \ref{fig1.1}, this controller is simpler than any controller except the Jacobian controller and outperforms any other controllers on trajectory following tasks.
In addition, such a controller demonstrates its adaptability in various scenarios, like different physical properties, frequencies, and even external disturbances. 
AdapJ achieves internal adaptability defined in \cite{24ZCc}, demonstrating adaptability across different robots.
\red{As a soft manipulator controller, this controller addresses hysteresis by including previous states and actuations as input and relaxing the parameter coupling, and addresses nonlinearity by updating online.}
As mentioned in \cite{24ZCa}, soft robot control researchers may blame the Jacobian controller's low efficiency and limited application on the linear format. 
However, this work demonstrates that a linear controller can achieve satisfying performance on soft robots. 
All we need is a proper linear format, not the one directly transferred from the rigid robot research.

Although the proposed AdapJ outperforms most controllers in terms of control accuracy, computational time, and adaptability, there are still limitations to this strategy.
First, theoretical guarantees remain underdeveloped.
The coverage of parameter updating leveraging the Gauss–Newton method has been proven in \cite{19ZT}.
However, providing formal proofs of coverage and robustness for data-driven controllers in soft robotics is still an open challenge.
A key difficulty lies in establishing an appropriate forward model that can adequately capture the complex dynamics of soft robots.
Although some studies have employed the Jacobian forward model for soft robotics \cite{24CP}, such a model is inherently limited due to the simplifying assumption ($\triangle s = J \triangle a$), which does not hold in soft robotic systems as discussed before.
Also, data-driven approaches, such as the statistical controllers \cite{19BY,19GF} and RNN controllers \cite{22DW, 23ZCc}, always meet challenges with theoretical validation due to the large parameter spaces and reliance on empirical training.
Although the parameters in our controller $A_*,B_*$ have been explained roughly in Sec. \ref{sec4.1.2}, providing better explainability than RNN, a rigorous theoretical analysis of their behavior and guarantees is still absent.
Enhancing the interpretability and provability of such adaptive data-driven controllers remains a vital direction for future work.
Second, this study primarily proposes this concept and initially validates its accuracy and adaptability in soft manipulators. 
In future work, we would like to extend and adapt this controller to more complex soft robots, such as modular soft robots and mobile soft robotic platforms.
The high dynamic behaviors of modular soft robots caused by the modularity may require more previous terms.
Also, the non-homogeneous matrix induced by the higher degrees of freedom may be a challenge.

\section*{Acknowledgement}
The authors would like to thank Godfried Jansen Van Vuuren for the help during robot setup manufacturing and Qinglin Zhu and Jingyi Sun for the discussion about the applications of neural networks.

\bibliographystyle{IEEEtran}
\bibliography{IEEEabrv,references}

\begin{thebibliography}{10}
\providecommand{\url}[1]{#1}
\csname url@samestyle\endcsname
\providecommand{\newblock}{\relax}
\providecommand{\bibinfo}[2]{#2}
\providecommand{\BIBentrySTDinterwordspacing}{\spaceskip=0pt\relax}
\providecommand{\BIBentryALTinterwordstretchfactor}{4}
\providecommand{\BIBentryALTinterwordspacing}{\spaceskip=\fontdimen2\font plus
\BIBentryALTinterwordstretchfactor\fontdimen3\font minus \fontdimen4\font\relax}
\providecommand{\BIBforeignlanguage}[2]{{%
\expandafter\ifx\csname l@#1\endcsname\relax
\typeout{** WARNING: IEEEtran.bst: No hyphenation pattern has been}%
\typeout{** loaded for the language `#1'. Using the pattern for}%
\typeout{** the default language instead.}%
\else
\language=\csname l@#1\endcsname
\fi
#2}}
\providecommand{\BIBdecl}{\relax}
\BIBdecl

\bibitem{11ST}
S.~Tully, G.~Kantor, M.~A. Zenati, and H.~Choset, ``Shape estimation for image-guided surgery with a highly articulated snake robot,'' in \emph{2011 IEEE/RSJ International Conference on Intelligent Robots and Systems}.\hskip 1em plus 0.5em minus 0.4em\relax IEEE, 2011, pp. 1353--1358.

\bibitem{14MY}
M.~C. Yip and D.~B. Camarillo, ``Model-less feedback control of continuum manipulators in constrained environments,'' \emph{IEEE Transactions on Robotics}, vol.~30, no.~4, pp. 880--889, 2014.

\bibitem{14JQ}
J.~F. Quei{\ss}er, K.~Neumann, M.~Rolf, R.~F. Reinhart, and J.~J. Steil, ``An active compliant control mode for interaction with a pneumatic soft robot,'' in \emph{2014 IEEE/RSJ International Conference on Intelligent Robots and Systems}.\hskip 1em plus 0.5em minus 0.4em\relax IEEE, 2014, pp. 573--579.

\bibitem{23ZT}
Z.~Tang, P.~Wang, W.~Xin, Z.~Xie, L.~Kan, M.~Mohanakrishnan, and C.~Laschi, ``Meta-learning-based optimal control for soft robotic manipulators to interact with unknown environments,'' in \emph{2023 IEEE International Conference on Robotics and Automation (ICRA)}.\hskip 1em plus 0.5em minus 0.4em\relax IEEE, 2023, pp. 982--988.

\bibitem{24ZCd}
Z.~Chen, Q.~Guan, J.~Hughes, A.~Menciassi, and C.~Stefanini, ``A versatile neural network configuration space planning and control strategy for modular soft robot arms,'' \emph{IEEE Transactions on Robotics}, 2025.

\bibitem{21HJ}
H.~Jiang, Z.~Wang, Y.~Jin, X.~Chen, P.~Li, Y.~Gan, S.~Lin, and X.~Chen, ``Hierarchical control of soft manipulators towards unstructured interactions,'' \emph{The International Journal of Robotics Research}, vol.~40, no.~1, pp. 411--434, 2021.

\bibitem{19ZT}
Z.~Q. Tang, H.~L. Heung, K.~Y. Tong, and Z.~Li, ``A novel iterative learning model predictive control method for soft bending actuators,'' in \emph{2019 International Conference on Robotics and Automation (ICRA)}.\hskip 1em plus 0.5em minus 0.4em\relax IEEE, 2019, pp. 4004--4010.

\bibitem{23ZCc}
Z.~Chen, X.~Ren, M.~Bernabei, V.~Mainardi, G.~Ciuti, and C.~Stefanini, ``A hybrid adaptive controller for soft robot interchangeability,'' \emph{IEEE Robotics and Automation Letters}, vol.~9, no.~1, pp. 875--882, 2023.

\bibitem{24XL}
X.~Li, Q.~Xiong, D.~Sui, Q.~Zhang, H.~Li, Z.~Wang, T.~Zheng, H.~Wang, J.~Zhao, and Y.~Zhu, ``Disturbance-adaptive tapered soft manipulator with precise motion controller for enhanced task performance,'' \emph{IEEE Transactions on Robotics}, vol.~40, pp. 3581--3601, 2024.

\bibitem{24ZCa}
Z.~Chen, F.~Renda, A.~Le~Gall, L.~Mocellin, M.~Bernabei, T.~Dangel, G.~Ciuti, M.~Cianchetti, and C.~Stefanini, ``Data-driven methods applied to soft robot modeling and control: A review,'' \emph{IEEE Transactions on Automation Science and Engineering}, vol.~22, no.~1, pp. 2241--2256, 2025.

\bibitem{18CS}
C.~Della~Santina, R.~K. Katzschmann, A.~Biechi, and D.~Rus, ``Dynamic control of soft robots interacting with the environment,'' in \emph{2018 IEEE International Conference on Soft Robotics (RoboSoft)}.\hskip 1em plus 0.5em minus 0.4em\relax IEEE, 2018, pp. 46--53.

\bibitem{18FR}
F.~Renda, F.~Boyer, J.~Dias, and L.~Seneviratne, ``Discrete cosserat approach for multisection soft manipulator dynamics,'' \emph{IEEE Transactions on Robotics}, vol.~34, no.~6, pp. 1518--1533, 2018.

\bibitem{19BY}
B.~Yu, J.~d.~G. Fern{\'a}ndez, and T.~Tan, ``Probabilistic kinematic model of a robotic catheter for 3d position control,'' \emph{Soft robotics}, vol.~6, no.~2, pp. 184--194, 2019.

\bibitem{19GF}
G.~Fang, X.~Wang, K.~Wang, K.-H. Lee, J.~D. Ho, H.-C. Fu, D.~K.~C. Fu, and K.-W. Kwok, ``Vision-based online learning kinematic control for soft robots using local gaussian process regression,'' \emph{IEEE Robotics and Automation Letters}, vol.~4, no.~2, pp. 1194--1201, 2019.

\bibitem{22DW}
D.~Wu, X.~T. Ha, Y.~Zhang, M.~Ourak, G.~Borghesan, K.~Niu, F.~Trauzettel, J.~Dankelman, A.~Menciassi, and E.~Vander~Poorten, ``Deep-learning-based compliant motion control of a pneumatically-driven robotic catheter,'' \emph{IEEE Robotics and Automation Letters}, vol.~7, no.~4, pp. 8853--8860, 2022.

\bibitem{17KLb}
K.-H. Lee, M.~C. Leong, M.~C. Chow, H.-C. Fu, W.~Luk, K.-Y. Sze, C.-K. Yeung, and K.-W. Kwok, ``Fem-based soft robotic control framework for intracavitary navigation,'' in \emph{2017 IEEE International Conference on Real-time Computing and Robotics (RCAR)}.\hskip 1em plus 0.5em minus 0.4em\relax IEEE, 2017, pp. 11--16.

\bibitem{18ML}
M.~Li, R.~Kang, D.~T. Branson, and J.~S. Dai, ``Model-free control for continuum robots based on an adaptive kalman filter,'' \emph{IEEE/ASME Transactions on Mechatronics}, vol.~23, no.~1, pp. 286--297, 2017.

\bibitem{17MY}
M.~C. Yip, J.~A. Sganga, and D.~B. Camarillo, ``Autonomous control of continuum robot manipulators for complex cardiac ablation tasks,'' \emph{Journal of Medical Robotics Research}, vol.~2, no.~01, p. 1750002, 2017.

\bibitem{17YJ}
Y.~Jin, Y.~Wang, X.~Chen, Z.~Wang, X.~Liu, H.~Jiang, and X.~Chen, ``Model-less feedback control for soft manipulators,'' in \emph{2017 IEEE/RSJ International Conference on Intelligent Robots and Systems (IROS)}.\hskip 1em plus 0.5em minus 0.4em\relax IEEE, 2017, pp. 2916--2922.

\bibitem{17TL}
T.~Liu, R.~Jackson, D.~Franson, N.~L. Poirot, R.~K. Criss, N.~Seiberlich, M.~A. Griswold, and M.~C. {\c{C}}avu{\c{s}}o{\u{g}}lu, ``Iterative jacobian-based inverse kinematics and open-loop control of an mri-guided magnetically actuated steerable catheter system,'' \emph{IEEE/ASME Transactions on Mechatronics}, vol.~22, no.~4, pp. 1765--1776, 2017.

\bibitem{17TTb}
T.~George~Thuruthel, E.~Falotico, M.~Manti, A.~Pratesi, M.~Cianchetti, and C.~Laschi, ``Learning closed loop kinematic controllers for continuum manipulators in unstructured environments,'' \emph{Soft robotics}, vol.~4, no.~3, pp. 285--296, 2017.

\bibitem{21GL}
G.~Li, T.~Stalin, P.~V. y~Alvarado \emph{et~al.}, ``Dnn-based predictive model for a batoid-inspired soft robot,'' \emph{IEEE Robotics and Automation Letters}, vol.~7, no.~2, pp. 1024--1031, 2021.

\bibitem{18TT}
T.~G. Thuruthel, E.~Falotico, M.~Manti, and C.~Laschi, ``Stable open loop control of soft robotic manipulators,'' \emph{IEEE Robotics and Automation Letters}, vol.~3, no.~2, pp. 1292--1298, 2018.

\bibitem{23ZCd}
Z.~Chen, M.~Bernabei, V.~Mainardi, X.~Ren, G.~Ciuti, and C.~Stefanini, ``A novel and accurate bilstm configuration controller for modular soft robots with module number adaptability,'' \emph{Soft robotics}, 2024.

\bibitem{20YW}
Y.-Y. Wu and N.~Tan, ``Model-less feedback control for soft manipulators with jacobian adaptation,'' in \emph{2020 International Symposium on Autonomous Systems (ISAS)}.\hskip 1em plus 0.5em minus 0.4em\relax IEEE, 2020, pp. 217--222.

\bibitem{24CP}
C.~Pan, Z.~Deng, C.~Zeng, B.~He, and J.~Zhang, ``Optimal visual control of tendon-sheath-driven continuum robots with robust jacobian estimation in confined environments,'' \emph{Mechatronics}, vol. 104, p. 103260, 2024.

\bibitem{24XR}
X.~Ren, T.~Pan, P.~Dario, S.~Wang, P.~W.~Y. Chiu, G.~Ciuti, and Z.~Li, ``Design and analytical modeling of a dumbbell-shaped balloon anchoring actuator for safe and efficient locomotion inside gastrointestinal tract,'' \emph{Soft Robotics}, 2024.

\bibitem{24ZCc}
Z.~Chen, D.~Wu, Q.~Guan, D.~Hardman, F.~Renda, J.~Hughes, T.~G. Thuruthel, C.~Della~Santina, B.~Mazzolai, H.~Zhao \emph{et~al.}, ``A survey on soft robot adaptability: Implementations, applications, and prospects [survey],'' \emph{IEEE Robotics \& Automation Magazine}, 2025.

\end{thebibliography}
\vfill
\begin{IEEEbiography}[{\includegraphics[width=1in,height=1.25in,clip,keepaspectratio]{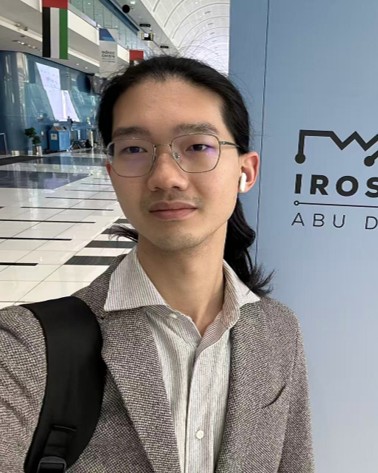}}]{Zixi Chen} received the M.Sc. degree in Control Systems from Imperial College in 2021. He is currently pursuing the Ph.D. degree in Biorobotics from Scuola Superiore Sant’Anna of Pisa.

His research interests include optical tactile sensors and data-driven approaches in soft robots.
\end{IEEEbiography}

\begin{IEEEbiography}[{\includegraphics[width=1in,height=1.25in,clip,keepaspectratio]{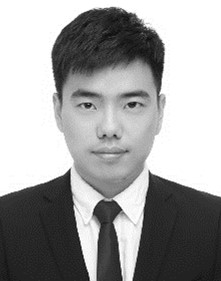}}]{Xuyang Ren} received the M.Sc. degree in mechanical engineering from Tianjin University, Tianjin, China, in 2019, and the Ph.D. degrees in biorobotics at the BioRobotics Institute of Scuola Superiore Sant’Anna, Pisa, Italy, in 2023. His main research interests include endoscopic devices, surgical robotics, and soft actuators for medical applications.
\end{IEEEbiography}

\begin{IEEEbiography}[{\includegraphics[width=1in,height=1.25in,clip,keepaspectratio]{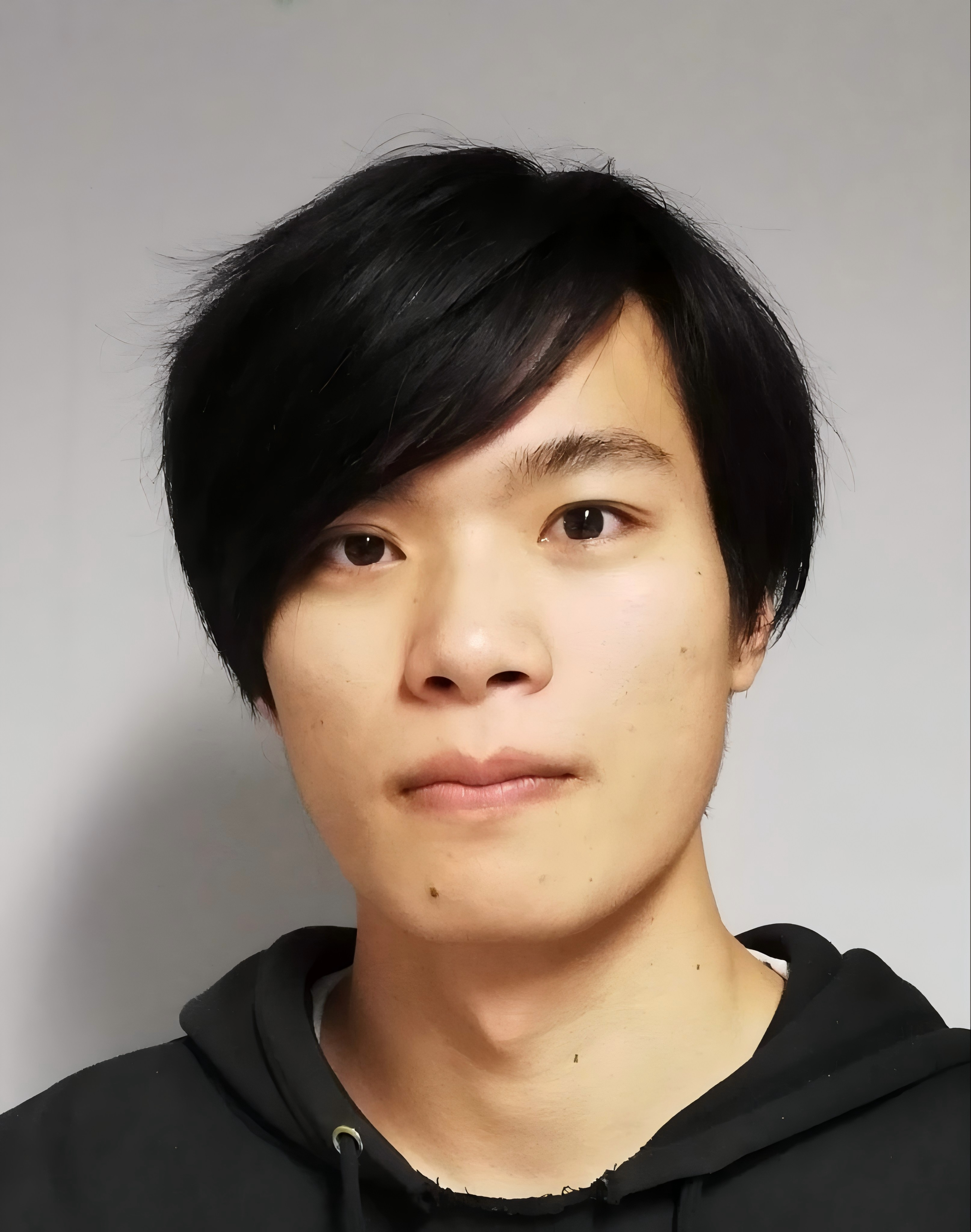}}]{Yuya Hamamatsu} received his M.Sc. degree in environment from the University of Tokyo, Japan, in 2020. He is currently working toward the Ph.D. degree with the Centre for Biorobotics at Tallinn University of Technology, Estonia. His research interests are control theory with machine learning techniques on field robotics.
\end{IEEEbiography}

\begin{IEEEbiography}[{\includegraphics[width=1in,height=1.25in, clip,keepaspectratio]{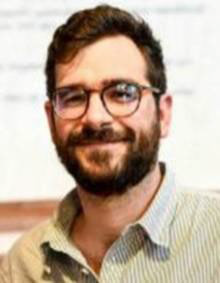}}]{Gastone Ciuti} (Senior Member, IEEE) received the master’s degree (Hons.) in biomedical engineering from the University of Pisa, Pisa, Italy, in 2008, and the Ph.D. degree (Hons.) in biorobotics from The BioRobotics Institute of Scuola Superiore Sant’Anna, Pisa, Italy, in 2011. He is currently an Associate Professor of Bioengineering at Scuola Superiore Sant’Anna, leading the Healthcare Mechatronics Laboratory. He has been a Visiting Professor at the Sorbonne University, Paris, France, and Beijing Institute of Technology, Beijing, China, and a Visiting Student at the Vanderbilt University, Nashville, TN, USA, and Imperial College London, London, U.K. He is the coauthor of more than 110 international peer reviewed papers on medical robotics and the inventor of more than 15 patents. His research interests include robot/computer-assisted platforms, such as tele-operated and autonomous magnetic-based robotic platforms for navigation, localization and tracking of smart and innovative devices in guided and targeted minimally invasive surgical and diagnostic applications, e.g. advanced capsule endoscopy. He is a Senior Member of the Institute of Electrical and Electronics Engineers (IEEE) society and Member of the Technical Committee in BioRobotics of the IEEE Engineering in Medicine and Biology Society (EMBS). He is an Associate Editor of the IEEE Journal of Bioengineering and Health Informatics, IEEE Transaction on Biomedical Engineering and of the IEEE Transaction on Medical Robotics and Bionics.
\end{IEEEbiography}

\begin{IEEEbiography}[{\includegraphics[width=1in,height=1.25in, clip,keepaspectratio]{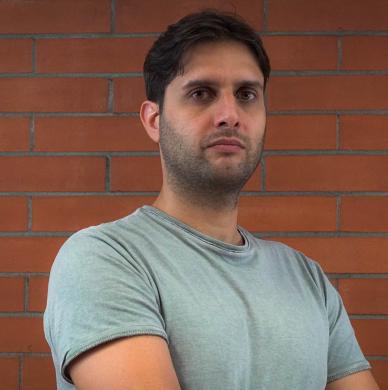}}]{Donato Romano} [M.Sc. in Agriculture, Food, and Environmental Science and Technologies (honors), PhD in BioRobotics (honors)] is currently Associate Professor at The BioRobotics Institute of Scuola Superiore Sant’Anna of Pisa, Italy, where he coordinates the Biorobotic Ecosystems Laboratory. Romano is mainly focusing his activities on bioinspired and biomimetic robotics, and in particular on animal-robot interaction, biohybrid systems, natural and biohybrid intelligence, ethorobotics, neuroethology. Romano received national and international recognitions for his research. He also has been visiting researcher at Khalifa University, Abu Dhabi (UAE). He is Member of the Editorial Board for many International Scientific Journals. Romano is Coordinator, PI, or partner of several national and international research projects.
\end{IEEEbiography}

\begin{IEEEbiography}[{\includegraphics[width=1in,height=1.25in,clip,keepaspectratio]{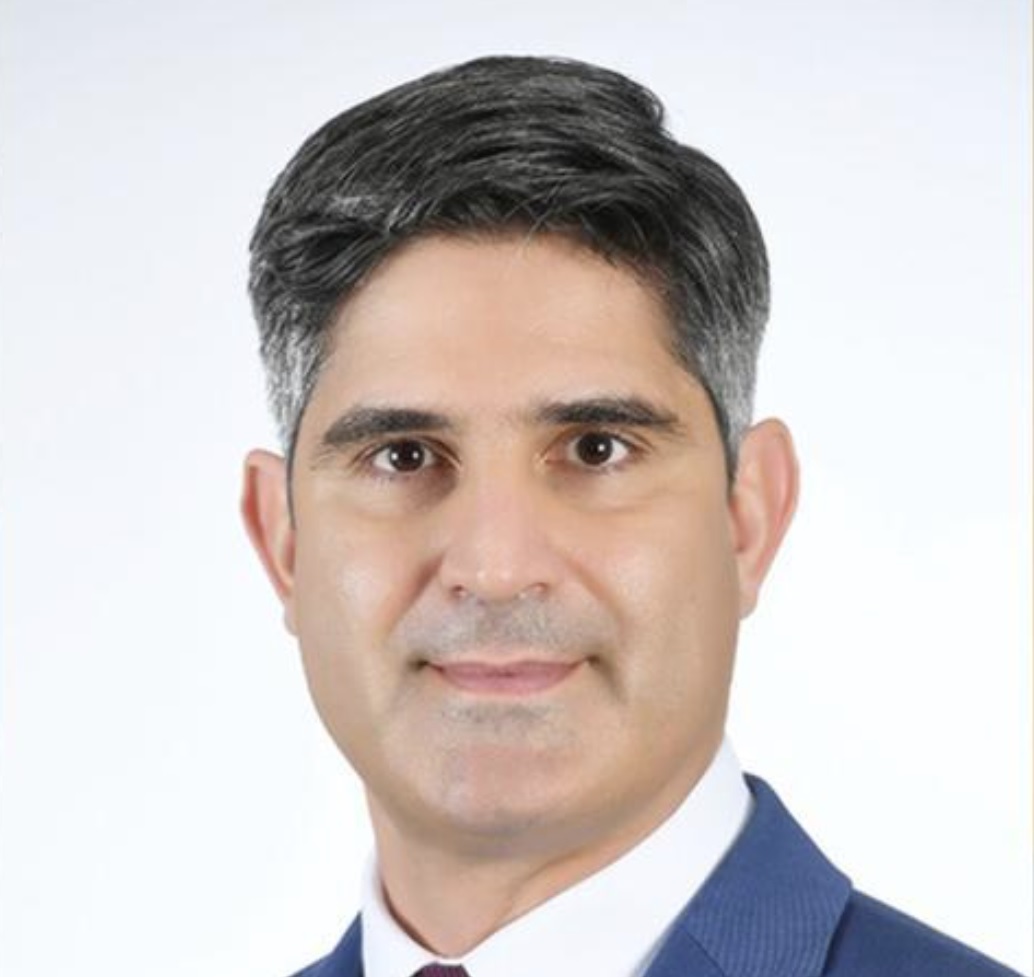}}]{Cesare Stefanini} (Member, IEEE) received the M.Sc. degree in mechanical engineering and the Ph.D. degree in microengineering, both with honors, from Scuola Superiore Sant-Anna (SSSA), Pisa, Italy, in 1997 and 2002, respectively.

He is currently Professor and Director of the BioRobotics Institute in the same University where he is also the Head of the Creative Engineering Lab. His research activity is applied to different fields, including underwater robotics, bioinspired systems, biomechatronics, and micromechatronics for medical and industrial applications. He received international recognitions for the development of novel actuators for microrobots and he has been visiting Researcher with the University of Stanford, Center for Design Research and the Director of the Healthcare Engineering Innovation Center, Khalifa University, Abu Dhabi, UAE.

Prof. Stefanini is the recipient of the “Intuitive Surgical Research Award.” He is the author or coauthor of more than 300 articles on refereed international journals and on international conferences proceedings. He is the inventor of more than 15 international patents, nine of which industrially exploited by international companies. He is a member of the Academy of Scientists of the UAE and of the IEEE Societies RAS (Robotics and Automation) and EMBS (Engineering in Medicine and Biology).
\end{IEEEbiography}
\end{document}